\documentclass[12pt]{article}

\usepackage{enumerate}

\usepackage{amsmath, amsthm, amssymb, bm, natbib,tikz}
\usetikzlibrary{arrows.meta}
\usepackage{framed,xcolor,geometry,setspace}
\usepackage{graphicx}
\usepackage{threeparttable,booktabs,lscape,pdflscape,multicol, url}

\usepackage{multirow,rotating, tabularx}
\usepackage{xcolor}

\usepackage{verbatim,epstopdf}
\usepackage{caption}
\usepackage{placeins}
\usepackage{fancybox,url}

\def\bSig\mathbf{\Sigma}

\def\iid{\overset{i.i.d.}{\sim}}
\newcommand{\blind}{0}

\newtheorem{proposition}{{\bf Proposition}}

\newtheorem{corollary}{{\bf Corollary}}
\newtheorem{assumption}{{\bf Assumption}}

\newtheorem{theorem}{{\bf Theorem}}

\newtheorem{example}{{\bf Example}}

\newtheorem{remark}{{\bf Remark}}

\def\E{\mathbb{E}}
\def\R{\mathbb R}
\def\cL{\mathcal L}
\def\hL{\hat{\mathcal L}}
\def\UnifNoise{{\rm Unif}([0,1]^q)}

\def\c{\mathcal}

\def\eps{\epsilon}

\def\ED{D_e}
\newcommand{\indep}{\;\, \rule[0em]{.03em}{.65em} \hspace{-.41em}
\rule[-.02em]{.65em}{.03em} \hspace{-.41em}
\rule[0em]{.03em}{.65em}\;\,}

\def\iid{\overset{i.i.d.}{\sim}}
\date{}
\begin{document}
\definecolor{shadecolor}{gray}{0.9}

\if0\blind
{
    \title{\Large Belted Engression: Sufficient Dimension Reduction for Generative Distributional Regression}
    \author{Wenxi Tan, Bing Li, and Lingzhou Xue \\ Department of Statistics, The Pennsylvania State University}
  \maketitle
} \fi

\if1\blind
{
  \title{\Large Belted Engression: Sufficient Dimension Reduction for Generative Distributional Regression}
  \maketitle
} \fi

\begin{abstract}
    Modern conditional generative models face significant challenges when learning complex covariate dependencies. While sufficient dimension reduction (SDR) provides a principled approach to compress these dependencies, traditional 
    SDR frameworks were not formulated for conditional generation. To bridge this gap, we propose Belted Engression, a unified and architecturally parameter-efficient framework for generative distributional regression. Our approach establishes an end-to-end compress-then-generate paradigm driven by sufficient representation learning, embedding a structural bottleneck into the generative architecture. Theoretically, we prove that the standard SDR condition is equivalent to a law-preserving generative factorization, which is achieved at the global optimum of the population Belted Engression objective. Furthermore, by uncovering a localized Bernstein-type control for the energy-score loss, we establish finite-sample convergence rates that are sharper than those of existing results. We also prove that this belted architecture is strictly smaller, operating with an asymptotically vanishing parameter count relative to the unstructured baseline. Extensive simulations and real-world applications demonstrate that Belted Engression achieves superior distributional prediction and SDR recovery with fewer trainable parameters.
\end{abstract}

\noindent{\it Keywords:} Energy Statistics; Generative Model; Neural Networks; Representation Learning.

\setstretch{1.5}

\section{Introduction}
Regression analysis seeks to understand the dependence of a response \(Y\) on covariates \(X\). While classical methods focus on point summaries, such as the conditional mean \(\E[Y|X=x]\) or conditional quantiles, a growing number of scientific applications require learning the full conditional distribution of \(Y\) given \(X=x\). Estimating this conditional law enables valid prediction intervals, conditional sampling, and the discovery of complex distributional effects beyond mean shifts. 
However, 
parametric families are often too rigid for modeling heteroscedasticity or multimodality in complex data, and traditional
nonparametric regression and density estimation scale poorly as the covariate dimension grows \citep{scott2015multivariate}.

To overcome these limitations, generative neural networks
\citep{kingma2013auto,goodfellow2014generative} have emerged as central tools for generating samples from complex, multivariate conditional laws. Among these generative approaches, Engression \citep{shen2025engression} has recently gained significant traction as a simple, flexible, and computationally lightweight framework for distributional regression. Trained via an energy-score-based objective, Engression models the conditional distribution by directly feeding both the full covariates $X$ and auxiliary random noise $\eta$ into a neural network generator. Engression remains empirically competitive and has found application in fields ranging from environmental
science
\citep{kraft2026modeling} to single-cell biology \citep{von2025representation}.

{
Despite recent advances in neural generative models, learning conditional laws in the presence of many covariates faces a significant statistical challenge. While Engression methods leverage latent noise to generate stochastic output variation, their generators remain conditioned on the full covariate vector $X\in\mathbb{R}^p$, and as demonstrated by \citet{huang2026theoretical} recently, this direct conditioning on an uncompressed covariate vector causes convergence rates to deteriorate sharply as the combined dimension of the input increases. Existing attempts to mitigate this dimension-driven complexity often rely on implicit structural assumptions, such as imposing hierarchical structures on the conditional quantile function \citep{chai2026neural} or assuming the conditional generator is composed of low-dimensional smooth functions \citep{chen2026error}. However, these theoretical assumptions describe the structure strictly through the functional form of the target generator, and they do not explicitly extract or isolate a separate low-dimensional summary of the response-relevant information in $X$. 

The success of modern deep learning relies heavily on extracting lower-dimensional, task-relevant features from complex data \citep{bengio2013representation}. 
Adopting this explicit representation learning perspective offers a more direct and interpretable alternative to implicit structural constraints. By isolating the low-dimensional structure into a distinct feature space, we can compress the covariates through a representational bottleneck before they enter the stochastic generation step. Image-valued covariates provide a compelling example: while an image may contain thousands of pixels with many reflecting noise or irrelevant background variation, its response-relevant features often lie on a lower-dimensional geometric structure \citep{roweis2000nonlinear}. Learning such a representation before conditional generation reduces both the effective covariate dimension and the complexity of the required generator. This potential advantage motivates a fundamental research question for distributional regression:

\emph{Can we leverage the principle of explicit representation learning \citep{bengio2013representation} to design an information-preserving representation for conditional distribution generation?}

This principle is statistically captured by sufficient dimension reduction 
{\citep[SDR;][]{li1991sliced,  CookWeisberg1991,cook1998regression, lee2013nonlinear,  li2018sufficient}},
  which assumes that there exists a mapping
\(s^*:\R^p\to\R^d\), with \(d<p\), such that
\[Y\indep X\mid s^*(X).\] 
The resulting sufficient predictor \(s^*(X)\)
retains all information in \(X\) relevant to the conditional law of \(Y\),
while providing an explicit and interpretable summary of the regression
relationship. However, integrating this representational structure into an energy-score-based conditional generative framework, such as Engression, is {by no means straightforward. Classical SDR estimation relies on inverse regression, forward regression, or reproducing kernel regression, which operate directly on the predictors, but in the context of Engression, there is an additional latent noise variable to be incorporated.}   Moreover, SDR seeks to
deterministically compress the covariates into low-dimensional summaries for inference, visualization, and prediction 
\citep{li2018sufficient,zhang2024Dimension,zhang2024nonlinear,xu2025conditional,tang2026belted}. In contrast, generative models
rely on expanding latent noise {to generate samples from the conditional distribution.} Hence, designing a single implicit generative objective that can simultaneously induce dimension reduction and accurate distributional sampling remains a critical gap.}

In this paper, we propose \textit{Belted Engression}, a novel and architecturally parameter-efficient generative framework for conditional distribution learning. To overcome the restrictive distributional assumptions of
inverse regression estimators and the inefficiencies of decoupled multistep pipelines, Belted Engression
establishes a \emph{compress-then-generate} paradigm that explicitly embeds
law-preserving dimension reduction into the generative architecture. Optimized end-to-end via a unified energy-score objective, it jointly
learns a sufficient-predictor mapping \(s\), termed the \textit{belt}, which acts as a sufficient representational bottleneck, alongside a reduced stochastic
generator, ensuring that the conditional generation is driven by a low-dimensional representation of \(X\). Thus, the learned belt provides an interpretable, low-dimensional summary of the dependence of \(X\) on \(Y\).

Equipped with this architecture, we establish the theoretical and empirical superiority of this unified approach. Our major contributions are summarized as follows.
\begin{itemize}
  \item \textbf{A unified compress-then-generate paradigm for supervised learning.} The architecture of Belted Engression redefines how covariates are handled in conditional generative modeling. While existing dimension-reduction generative models such as the variational autoencoders (VAEs) \citep{kingma2013auto} and the distributional principal autoencoder (DPA) \citep{shen2024distributional} are designed for unsupervised representation learning, Belted Engression provides a flexible and lightweight end-to-end \textit{supervised} framework. By embedding a sufficient representational bottleneck into the energy-score-based objective, our approach structurally offers an interpretable and architecturally parameter-efficient solution for complex distribution learning.

  \item \textbf{A novel theoretical bridge between SDR and generative factorization.} We establish a key bridge between classical SDR and modern deep generative architectures. Specifically, we prove that the standard SDR condition is mathematically equivalent to a \textit{compress-then-generate} factorization, wherein a low-dimensional sufficient predictor combined with auxiliary noise is sufficient to reproduce the full conditional law. Through Propositions~\ref{prop:sufficient-predictor-randomization}--\ref{prop:minLsh}, we establish that this law-preserving factorization is achieved at the global optimum of the population-level Belted Engression objective, providing a definitive theoretical justification for our architectural design. This proves that our framework recovers the true conditional law, matching the theoretical target of standard Engression while performing dimension reduction.

    \item \textbf{Refined finite-sample theory via localized complexity control.}
    We develop a finite-sample analysis of Engression under standard
    H\"older smoothness conditions. Our proof exploits a localized Bernstein-type
    control of the Engression loss (see Remark~\ref{rem:mn bd for Bengression}),
    which yields a tighter bound for near-optimal candidates
    \citep{bartlett2006empirical}. We use this strategy to establish the
    finite-sample convergence rate for Belted Engression in
    Theorem~\ref{thm:BeltedEngression-rate}. Applied to Engression, it yields
    the refined rate in Proposition~\ref{prop:Engression-rate}, which improves upon
    the bounds of \citet{huang2026theoretical} and \citet{chen2026error}.

  \item \textbf{Provable statistical and architectural efficiency.} 
  We establish that Belted Engression improves on unstructured direct Engression in both statistical accuracy and model complexity.
  Theorem~\ref{thm:rate-comparison} demonstrates that our approach yields a finite-sample convergence rate that is sharper than the Engression counterpart, while Theorem~\ref{thm:size-comparison} proves that this belted architecture is strictly smaller, operating with an asymptotically vanishing parameter count relative to the unstructured baseline. These theoretical guarantees are supported by our simulations and real-world applications, confirming that Belted Engression simultaneously achieves accurate conditional distribution learning, SDR recovery, and parameter efficiency to deliver superior conditional distribution learning while utilizing smaller, lighter models.
\end{itemize}

The rest of the paper is organized as follows. Section~\ref{sec:preliminary}
reviews the background and establishes the necessary notation. Section~\ref{sec:BeltedEngression}
introduces the proposed Belted Engression framework.
Section~\ref{sec:theory} establishes our main theoretical results, developing the localized finite-sample convergence rates and architectural parameter efficiency bounds. Sections~\ref{sec:Simulation}
 and~\ref{sec:rda} demonstrate the empirical advantages of our approach through simulations and real-world applications, respectively. Section \ref{sec:discussion} includes a few concluding remarks. All proofs and additional technical details are deferred to the Supplementary Materials.

\section{Preliminaries}\label{sec:preliminary}

Engression \citep{shen2025engression} models the conditional distribution of a response given covariates via a generative process. It transforms a latent random vector with a known reference distribution, given covariates, into samples from the conditional law of the response.

\def\hi#1{^{#1}}
\def\real{\mathbb{R}}
\def\lo#1{_{#1}}
\def\ca#1{{\cal #1}}
\def\inv{\hi {-1}}
\def\of{\circ}

{\textbf{The generator map.} Let $(\Omega, \ca F, P)$ be a probability space, and $(\Omega \lo X, \ca F \lo X)$, $(\Omega \lo Y, \ca F \lo Y)$,   $(\Omega \lo \eta, \ca F \lo \eta)$ be measurable spaces, where $\Omega \lo X \subseteq \real \hi p$, $\Omega \lo Y \subseteq \real \hi r$, and $\Omega \lo \eta \subseteq \real \hi q$. Let $X: \Omega \rightarrow \Omega \lo X$ and $Y: \Omega \rightarrow \Omega \lo Y$ be random vectors measurable with respect to $\ca F / \ca F \lo X$ and $\ca F / \ca F \lo Y$, respectively.  Let \(P_X = P \circ X \inv \) and $P \lo Y = P \circ Y \inv$ be the distributions of $X$ and $Y$, respectively, and \(P_{Y\mid X}\) is the conditional distribution of \(Y\) given \(X\). Let $\eta: \Omega \to \Omega \lo \eta$ be a latent {random} vector independent of $(X,Y)$, with distribution $P_{\eta}= P \of \eta \inv$, measurable with respect to $\ca F / \ca F \lo \eta$.   }

{The objective of Engression is to learn a generator map \(g^*: \Omega \lo X  \times \Omega \lo \eta \rightarrow \Omega \lo Y\) such that  {the conditional distribution of $g \hi *(X, \eta)$ given $X$  } matches the conditional law of $Y$ given $X$. For any fixed \(x\in \Omega \lo X \), write the conditional law of \(Y\) given \(X=x\) as
\(P_{Y\mid X}(\cdot\mid x)\). For any generator map {\(g: \Omega \lo X  \times \Omega \lo \eta \to \Omega \lo Y\)}, let \(P_g(\cdot\mid x)\) denote the {conditional} law of \(g(x,\eta)\) {given $X$}  evaluated at a fixed \(x\). Thus, the {goal of Engression} is to find a generator {map $g \hi *$} satisfying
\begin{equation}\label{eq:Engression}
P_{g^*}(\cdot\mid x)=P_{Y\mid X}(\cdot\mid x)
\text{ for }P_X\text{-almost every  }x.
\end{equation}
Once $g^*$ is learned, one can draw samples from the conditional distribution $P_{Y\mid X}(\cdot\mid x)$ by generating \(\eta\sim P_{\eta}\) and evaluating \(g^*(x,\eta)\).  }

\def\argmin{\mathrm{argmin}}

\textbf{The energy loss.} {Let $L \lo 1 ( \Omega \lo X \times \Omega \lo \eta, \Omega \lo Y)$ be the class of all functions $g: \Omega \lo X \times \Omega \lo \eta \to \Omega \lo Y$ that are measurable with respect to $\ca F \lo X \times \ca F \lo \eta/ \ca F \lo Y$ such that
$    E \| g ( X, \eta ) \| < \infty. $ We assume that $g \hi * \in L \lo 1 (\Omega \lo X \times \Omega \lo \eta, \Omega \lo Y)$ and $\E \| Y \| < \infty$. For a probability measure $\mu$ on $(\Omega \lo Y, \ca F \lo Y)$ and $y \in \Omega \lo Y$,  let  $\mathrm{ES}(\mu,y) = \E  \|y-Z \| -\frac12\E \|Z-Z'\| $ where $Z, Z' \overset{iid}{\sim} \mu$. This quantity   is called the energy score. 
  Engression  leverages the energy score \(\operatorname{ES}(P \lo g (\cdot| x) ,y) \) to avoid the need for restrictive likelihood assumptions. Specifically, let $\cL(g) = \E [ \mathrm{ES} ( P \lo g (\cdot|X), Y)]$. 
By $\eta \indep (X,Y)$, $\cL (g)$ can be rewritten as 
\begin{equation}\label{eq:population-Engression-risk}
\cL(g) = \E\left[
\|Y-g(X,\eta)\|
-\frac12\|g(X,\eta)-g(X,\eta')\|
\right].
\end{equation}
\citet[Proposition 1]{shen2025engression} shows that \(g^*\) in \eqref{eq:Engression} is the minimizer of $\ca L (g)$:
  {$g^*\in\argmin_{g \in L \lo 1 (\Omega \lo X \times \Omega \lo \eta, \Omega \lo Y)}\cL(g)$. }
Hence, the energy loss provides a principled approach to learning the conditional law of \(Y\) given \(X\) via the empirical risk minimization of \(\cL(g)\). }

\textbf{Neural network generators.}
Neural networks are central to this framework, as they can approximate complex, nonlinear generators and produce conditional samples through a single forward pass without requiring explicit density estimation. Formally, we parameterize these generators using feed-forward ReLU neural networks. 
A feed-forward network with
input dimension \(a\), output dimension \(b\), and \(D\) hidden layers is represented as a function $f:\R^a\to\R^b$ satisfying
\(
f(x)=L_D\circ \sigma\circ L_{D-1}\circ \sigma\circ \cdots\circ\sigma\circ L_0(x),
\)
where \(L_\ell(u)=A_\ell u+b_\ell\) for $\ell = 0, \dots D$, and \(\sigma(t)=\max\{t,0\}\) is the ReLU activation function applied componentwise. 

For each $\ell = 1, \dots D$, the function $ L_\ell \circ \sigma $ constitutes the $\ell$-th hidden layer, with $A_\ell\in \R^{p_{\ell+1}\times p_{\ell}}$ as the weight matrix and $b_\ell\in \R^{p_{\ell+1}}$ as the bias vector. Stacking the layer widths as
\((p_0,p_1,\ldots,p_D,p_{D+1})^\top\), with \(p_0=a\) and \(p_{D+1}=b\), the
network width is defined as \(W=\max_{1\le \ell\le D}p_\ell\).  Let $S$ denote the total number of affine parameters (i.e., network size), and for fully connected layers,
the parameter count is \(\sum_{\ell=0}^{D}p_{\ell+1}(p_\ell+1)\). 

While the 
parameters \(W,D\) and $S$ may depend on the sample size \(n\), we suppress this dependence in the notation for brevity. Assuming, without loss of generality, that the inputs are bounded in the unit hypercube $[0,1]^a$, we let \(\c F_{NN}^{a,b}(W,D,S,\bar B)\) denote the corresponding ReLU network class
with input dimension \(a\), output dimension \(b\), width \(W\), depth \(D\),
size \(S\), and a uniform envelope bound \(\bar B>0\) such that
\(
\max_{1\le j\le b}\sup_{u\in[0,1]^a}|f_j(u)|\le \bar B .
\)

\section{Methodology}\label{sec:BeltedEngression}
This section formalizes the Belted Engression framework in two stages. First, we establish that the
SDR condition induces a structured, factorized generative representation that preserves the target
conditional law. Building upon this insight, we then construct the
Belted Engression objective {function} and introduce its corresponding neural network estimator.

\subsection{SDR Factorization}
In regression problems, the information in   \(X\in\R^p\)  about $Y$ can often be captured by a lower-dimensional representation, such as indices, latent factors, or nonlinear representations;  see, for example, \cite{li1991sliced},  \cite{li2018sufficient}, and \cite{fan2017sufficient}. 
This is the premise of  sufficient dimension reduction (SDR), which  assumes the existence of a low-dimensional function \(s^*(X)\) such that \(Y\indep X\mid s^*(X).\) Thus, \(s^*(X)\), referred to as a sufficient predictor, preserves all information in \(X\) that is relevant to modeling \(Y\). 

The following proposition formalizes a critical theoretical bridge between SDR and generative factorization. Specifically, we will show that the SDR condition is equivalent to the existence of a law-preserving factorization of the conditional law \(P_{Y\mid X}\) through a compress-then-generate approach.  {Let $\Omega \lo S$ be a measurable set in $\real \hi d$}. For measurable maps {\(s:\Omega \lo X \to \Omega \lo S\) and \(h:\Omega \lo S \times \Omega \lo \eta \to \Omega \lo Y\)}, let \(P_{s,h}(\cdot\mid X)\) denote the conditional law of the generated output \(h(s(X),\eta)\) given \(X\), and we define
\(P_{s,h}(\cdot\mid x)\) as the law of \(h(s(x),\eta)\) with a fixed \(x\).
\begin{proposition}\label{prop:sufficient-predictor-randomization}
{Let $\Omega \lo \eta = [0, 1] \hi q$.} The following statements are equivalent.

(i) There exists a measurable map \(s^*:\Omega \lo X \to \Omega \lo S \)
such that \(X\indep Y \mid s^*(X)\).

\def\UnifNoise{\mathrm{Unif}(\Omega \lo \eta)}

(ii) There exist measurable maps \(s^*:\Omega \lo X \to \Omega \lo S \) and
\(h^*:\Omega \lo S \times \Omega \lo \eta \to \Omega \lo Y\) such that, with \(\eta\sim {\UnifNoise}\)
independent of {\((X, Y)\)},
\begin{equation}\label{eq:sufficient-predictor-generator}
P_{s^*,h^*}(\cdot\mid x)=P_{Y\mid X}(\cdot\mid x)
\quad\text{for }P_X\text{-almost every }x .
\end{equation}
\end{proposition}
The factorization {described} in
Proposition~\ref{prop:sufficient-predictor-randomization} is 
inherently non-unique. For example, any bijective transformation of the sufficient predictor \(s^*\), together with
an appropriate modification of the generator \(h^*\), yields the identical conditional law.
Our theoretical analysis therefore focuses on the conditional law induced by the
factorization rather than the specific functional forms of \(s^*\) and \(h^*\). While 
Proposition~\ref{prop:sufficient-predictor-randomization} is stated using uniform noise to facilitate the subsequent neural network approximation theory, this representation holds analogously for any
atomless noise distribution $P_\eta$, including the standard Gaussian distribution.

From a representation learning perspective \citep{bengio2013representation}, the map \(s^*\) acts as a \emph{sufficient representational bottleneck}. It structurally compresses the $p$-dimensional covariate space into a $d$-dimensional feature while guaranteeing no loss of generative capacity for $Y$.

To facilitate the estimation of these components via neural networks, we assume that \(s^*(X)\) can be standardized to \([0,1]^d\). {That is, we take $\Omega \lo S$ to be $[0,1] \hi d$.} Because any bounded sufficient predictor can be mapped into \([0,1]^d\) via an affine coordinate-wise rescaling without violating
the conditional independence \(X\indep Y\mid s^*(X)\), this requirement imposes no practical restriction. We formally state this compact representation as our structural assumption.
\begin{assumption}\label{as:sdr-borel}
 There exist an integer \(d<p\) and a measurable map
\(s^*:{\Omega \lo X }\to [0,1]^d={\Omega \lo S}\) such that \(X\indep Y \mid s^*(X)\).
\end{assumption}
\noindent Under Assumption~\ref{as:sdr-borel},
Proposition~\ref{prop:sufficient-predictor-randomization}(ii) shows the existence of a measurable map
{\(h^*:\Omega \lo S \times \Omega \lo \eta \to \Omega \lo Y \)} satisfying
\eqref{eq:sufficient-predictor-generator}. Throughout, we fix
one such pair \((s^*,h^*)\) as the population reference.

\subsection{Population Objective and Empirical Estimator}
Building upon the law-preserving factorization established in Proposition~\ref{prop:sufficient-predictor-randomization}, we introduce the core architecture of Belted Engression. Elevating the ``belted'' network topologies of \citet{tang2026belted} to the generative setting, we represent the conditional law of \(Y\) given \(X\) via a composite generator \(h^*(s^*(X),\eta)\). In this framework, \(s^*:{\Omega \lo X \to \Omega \lo S} \)  serves as the \textit{belt}{---a sufficient representational bottleneck---}that explicitly compresses the ambient covariates \citep{bengio2013representation}. This low-dimensional feature manifold is then fed into the stochastic \textit{generator} \(h^*:{\Omega \lo S \times \Omega \lo \eta \to \Omega \lo Y}\). We now show that the pair \((s^*,h^*)\) can be characterized as the population risk minimizer of a structured energy loss.

Formally, we define the population objective of Belted Engression as
\begin{equation*}
\cL(s,h)
=
\E\left[
\|Y-h(s(X),\eta)\|
-\frac12\|h(s(X),\eta)-h(s(X),\eta')\|
\right], 
\end{equation*}
where {$\eta, \eta ' \overset{iid}\sim P \lo \eta$ and $(\eta, \eta') \indep (X,Y)$. }
{A pair \((s,h)\) is {said to be}  \emph{admissible} if its components are measurable maps satisfying
\(\E[\|h(s(X),\eta)\|]<\infty\).}

This objective \(\cL(s,h)\) connects to the target conditional law \(P_{Y\mid X}\) via the energy distance \citep{szekely2013energy}.
Recall that for two probability laws \(P,Q\) defined on the same Euclidean space, the squared energy distance between $P$ and $Q$ is given by
\[
\ED^2(P,Q)
=2\E[\|A-B\|]-\E[\|A-A'\|]-\E[\|B-B'\|],
\]
where \(A,A'\iid P\) and \(B,B'\iid Q\) are independent variables. Under Assumption~\ref{as:sdr-borel}, the above distances yield a crucial identity for the excess risk of any admissible pair \((s,h)\):
\[
\cL(s,h)-\cL(s^*,h^*)
=
\frac12 {\E }\left[\ED^2\left(P_{Y\mid X}(\cdot\mid X),P_{s,h}(\cdot\mid X)\right)\right] \ge 0,
\]
{where $\E$ is the expectation with respect to   \(X\)}. By the properties of the energy distance, equality holds if and only if \(P_{s,h}(\cdot\mid X)=P_{Y\mid X}(\cdot\mid X)\)  {almost surely $P$}; see Supplementary Lemma~S.1 for the proof. This identity yields the following population-risk characterization.
\begin{proposition}\label{prop:minLsh}
Under Assumption~\ref{as:sdr-borel} and assuming $\E[\|Y\|]<\infty$,
\begin{enumerate}[(i)]
\setlength{\itemsep}{0pt}
\setlength{\parsep}{0pt}
\setlength{\topsep}{0pt}
\item \((s^*,h^*)\) minimizes \(\cL(s,h)\) over all admissible measurable pairs \((s,h)\).
\item If \(g^*\) satisfies \eqref{eq:Engression}, then
{\(P_{s^*,h^*}(\cdot\mid X)=P_{g^*}(\cdot\mid X)=P_{Y\mid X}(\cdot\mid X)\) almost surely $P$.}
\end{enumerate}
\end{proposition}
Proposition~\ref{prop:minLsh} shows that our SDR factorization is a valid
population minimizer that recovers the target conditional law as unstructured Engression.
Thus, embedding the belt reduces the parameterization of the generator without compromising its generative capacity.

\begin{figure}[!ht]
\centering
\resizebox{0.90\textwidth}{!}{\begin{tikzpicture}[
  neuron/.style={circle, draw=black!42, fill=white, minimum size=2.8mm, inner sep=0pt},
  input/.style={circle, draw=cyan!55!black, fill=cyan!13, minimum size=3.8mm, inner sep=0pt, font=\tiny},
  latent/.style={circle, draw=cyan!45!black, fill=cyan!8, minimum size=4.3mm, inner sep=0pt, font=\tiny},
  noise/.style={circle, draw=black!45, fill=gray!14, minimum size=4.2mm, inner sep=0pt, font=\tiny},
  output/.style={circle, draw=orange!60!black, fill=orange!23, minimum size=3.8mm, inner sep=0pt, font=\tiny},
  edge/.style={draw=black!23, -{Latex[length=0.8mm,width=0.5mm]}, line width=0.13pt},
  interfaceedge/.style={draw=black!32, -{Latex[length=0.9mm,width=0.58mm]}, line width=0.18pt},
  sedge/.style={draw=blue!23, -{Latex[length=0.8mm,width=0.5mm]}, line width=0.14pt},
  sinterfaceedge/.style={draw=blue!23, -{Latex[length=0.9mm,width=0.58mm]}, line width=0.18pt},
  hedge/.style={draw=orange!42, -{Latex[length=0.8mm,width=0.5mm]}, line width=0.14pt},
  hinterfaceedge/.style={draw=orange!48, -{Latex[length=0.9mm,width=0.58mm]}, line width=0.18pt},
  inputarrow/.style={draw=cyan!45, -{Latex[length=1mm,width=0.65mm]}, line width=0.18pt},
  noisearrow/.style={draw=black!38, -{Latex[length=1mm,width=0.65mm]}, line width=0.18pt},
  lawarrow/.style={draw=orange!48, double distance=3pt, -{Implies}, line width=0.24pt},
  lawpanel/.style={rounded corners=2pt, draw=orange!42, fill=orange!4, line width=0.28pt, font=\tiny, align=center, inner xsep=4pt, inner ysep=3pt},
  partbox/.style={rounded corners=2.2pt, line width=0.28pt},
  boxtitle/.style={font=\tiny, inner sep=1.2pt},
  layerlabel/.style={font=\tiny, text=black!62, inner sep=1pt},
  interfacelabel/.style={font=\tiny, fill=white, fill opacity=0.9, text opacity=1, inner sep=1.1pt, align=center}
]
  \filldraw[partbox, draw=cyan!38, fill=cyan!9] (-2.40,-0.44) rectangle (-0.88,0.44);
  \filldraw[partbox, draw=cyan!45, fill=cyan!10, dashed] (-0.40,-1.1) rectangle (0.40,1.1);
  \filldraw[partbox, draw=orange!48, fill=orange!12, dashed] (5.74,-0.67) rectangle (6.50,0.55);
  \node[font=\tiny, align=center, inner sep=1.2pt] at (-1.6,0) {\shortstack{\(x\in\R^p\)\\(Predictors)}};
  \draw[inputarrow] (-0.88,0) -- (-0.40,0);
  \node[font=\tiny, text=blue!55!black, inner sep=1.2pt] at (1.10,1.6) {\shortstack{Belt \\ \(s:\R^p\to\R^d\)}};
  \node[font=\tiny, text=orange!62!black, inner sep=1.2pt] at (4.92,1.6) {\shortstack{Generator \\ $h: \R^d \times \R^q \to \R^r$}};

  \foreach \i/\y in {1/0.78,2/0.26,3/-0.26,4/-0.78}
    \node[input] (x\i) at (0,\y) {\(x_{\i}\)};
  \foreach \i/\y in {1/0.98,2/0.59,3/0.2,4/-0.2,5/-0.59,6/-0.98}
    \node[neuron, draw=blue!23, fill=blue!7] (a\i) at (1.05,\y) {};
  \foreach \i/\y in {1/0.98,2/0.59,3/0.2,4/-0.2,5/-0.59,6/-0.98}
    \node[neuron, draw=blue!23, fill=blue!7] (b\i) at (2.05,\y) {};
  \foreach \i/\y in {1/0.25,2/-0.25}
    \node[latent] (z\i) at (3.26,\y) {\(z_{\i}\)};
  \filldraw[partbox, draw=black!35, fill=gray!8, dashed] (2.94,-1.14) rectangle (3.58,-0.50);
  \filldraw[partbox, draw=black!32, fill=gray!8] (2.58,-2.16) rectangle (3.94,-1.56);
  \node[noise] (e) at (3.26,-0.82) {\(\eta\)};
  \node[font=\tiny, align=center, inner sep=1.2pt] at (3.26,-1.86) {\shortstack{\(\eta\sim P_\eta\)\\(Noise)}};
  \draw[noisearrow] (3.26,-1.56) -- (3.26,-1.14);
  \foreach \i/\y in {1/0.7,2/0.23,3/-0.23,4/-0.7}
    \node[neuron, draw=orange!58, fill=orange!9] (c\i) at (4.42,\y) {};
  \foreach \i/\y in {1/0.7,2/0.23,3/-0.23,4/-0.7}
    \node[neuron, draw=orange!58, fill=orange!9] (d\i) at (5.25,\y) {};
  \foreach \i/\y in {1/0.22,2/-0.36}
    \node[output] (y\i) at (6.12,\y) {\(y_{\i}\)};

  \foreach \i in {1,...,4}
    \foreach \j in {1,...,6}
      \draw[sedge] (x\i) -- (a\j);
  \foreach \i in {1,...,6}
    \foreach \j in {1,...,6}
      \draw[sedge] (a\i) -- (b\j);
  \foreach \i in {1,...,6}
    \foreach \j in {1,2}
      \draw[sinterfaceedge] (b\i) -- (z\j);
  \foreach \i in {1,2}
    \foreach \j in {1,...,4}
      \draw[hinterfaceedge] (z\i) -- (c\j);
  \foreach \j in {1,...,4}
    \draw[hinterfaceedge] (e) -- (c\j);
  \foreach \i in {1,...,4}
    \foreach \j in {1,...,4}
      \draw[hedge] (c\i) -- (d\j);
  \foreach \i in {1,...,4}
    \foreach \j in {1,2}
      \draw[hedge] (d\i) -- (y\j);

  \draw[lawarrow] (6.7,-0.07) -- (7.55,-0.07);
  \node[lawpanel] at (8.7,-0.07)
  {\shortstack{Conditional law \\of $Y\mid X=x$\\
  \(\approx h(s(x),\cdot)_{\#}P_\eta\)}};
  \draw[lawarrow] (8.7,-0.6) -- (8.7,-1);
  \node[lawpanel] at (8.7,-1.6)
  {\shortstack{Conditional generation\\  $\{h(s(x),\eta_j)\}_{j=1}^m$\\
   via \(\eta_1, \dots, \eta_m \sim P_\eta\)}};
\end{tikzpicture}
}
\caption{Belted Engression architecture for generative distributional regression, illustrated with \(p=4\), \(d=2\), \(q=1\), and \(r=2\).
The \textit{belt} maps the covariates \(X\) into a sufficient low-dimensional representation \(s(X)\), and the generator combines \(s(X)\) with simulation noise to approximate the conditional law of \(Y\) given $X=x$.
Network depths and widths for this schematic are \(W_s=6\), \(D_s=2\) for the belt, and \(W_h=4\), \(D_h=2\) for the generator.}
\label{fig:BeltedEngression-backbone}
\end{figure}
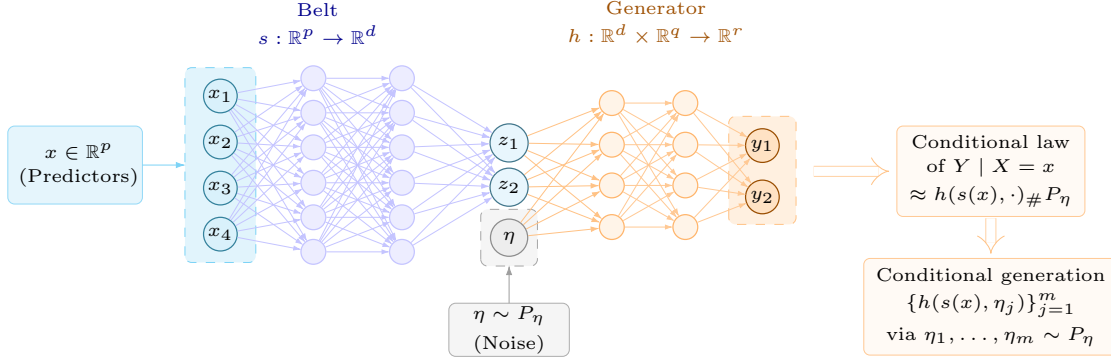

To construct a practical estimator, we parameterize the belt \(s\) and the generator \(h\) using the ReLU neural network classes introduced
in Section~\ref{sec:preliminary}. Specifically, we consider the spaces
\(
s\in\c S_{NN}\) and \(
h\in\c H_{NN},
\)
where
$\c S_{NN}=
\{s\in\c F_{NN}^{p,d}(W_s,D_s,S_s,\bar B_s):
s([0,1]^p)\subseteq[0,1]^d\}$,
and
$\c H_{NN}=\c F_{NN}^{d+q,r}(W_h,D_h,S_h,\bar B_h)$. The network parameters will be specified in Section~\ref{sec:theory}. Visually and functionally, this parameterization structurally induces the sufficient representational bottleneck, as illustrated in Figure~\ref{fig:BeltedEngression-backbone}.

Given an i.i.d. sample \(\{(X_i,Y_i)\}_{i=1}^n\) drawn from \((X,Y)\), we approximate the population objective by drawing
 \(m\ge2\) independent simulation noise vectors
\(\eta_{i1},\ldots,\eta_{im}\) from
\(P_\eta=\operatorname{Unif}([0,1]^q)\) for each observation $i$. This yields the empirical objective of Belted Engression:
\begin{equation*}
\hL(s,h)=
\frac1n\sum_{i=1}^n
\left[
\frac1m\sum_{j=1}^m
\|Y_i-h(s(X_i),\eta_{ij})\|
-
\frac{1}{2m(m-1)}
\sum_{j\ne j'}
\|h(s(X_i),\eta_{ij})-h(s(X_i),\eta_{ij'})\|
\right].
\end{equation*}
The Belted Engression estimator $(\hat s,\hat h)$ is defined as the empirical risk minimizer of this objective over the neural
network classes \(\c S_{NN}\) and \(\c H_{NN}\):
\begin{equation*}
(\hat s,\hat h)\in\arg\min_{s\in\c S_{NN},\;h\in\c H_{NN}}
\hL(s,h).
\end{equation*}
Once trained, generating synthetic samples for a new covariate observation $x$ is highly efficient. By drawing $R$ 
independent noise vectors \(\eta_1,\ldots,\eta_R\),  a single forward pass computes the generated samples \(\hat h(\hat s(x),\eta_1),\ldots,\hat h(\hat s(x),\eta_R)\). These simulated draws can then be used to estimate quantities such as conditional means, quantiles, and prediction intervals.

\section{Theoretical Properties}\label{sec:theory}

This section establishes the finite-sample theoretical guarantees for Belted Engression and its unstructured Engression baseline. We first derive the convergence rates for both estimators under comparable
H\"older smoothness and network capacity conditions. By contrasting the resulting rate exponents and the parameter counts of the prescribed architectures, we will demonstrate how embedding the sufficient representational bottleneck simultaneously improves statistical efficiency and reduces overall architectural complexity.

\subsection{Convergence Rates for Belted Engression}
We introduce the regularity conditions required to establish the convergence rates.
\begin{assumption}\label{as:bounded}
Assume \(X\in[0,1]^p\) and \(\|Y\|\le B_Y\) almost surely for some
\(B_Y>0\).
\end{assumption}
This boundedness condition is a standard regularity assumption in the theoretical analyses; see
\cite{song2026wasserstein,huang2026theoretical, chai2026neural}.

Next, let \(\beta>0\), and let \(\lfloor\beta\rfloor\) denote the largest integer
strictly smaller than \(\beta\). For a positive integer \(k\) and \(B_0>0\), we
define the H\"older class as
\begin{align*}
&\c H^\beta([0,1]^k,B_0) \\
&\quad =
\Big\{f:[0,1]^k\to\R:
\max_{\substack{\alpha\in\mathbb N_0^k\\ \|\alpha\|_1\le \lfloor\beta\rfloor}}
\|\partial^\alpha f\|_\infty\le B_0,\
\max_{\substack{\alpha\in\mathbb N_0^k\\ \|\alpha\|_1=\lfloor\beta\rfloor}}
\sup_{\substack{x,y\in[0,1]^k\\ x\ne y}}
\frac{|\partial^\alpha f(x)-\partial^\alpha f(y)|}
{\|x-y\|^{\beta-\lfloor\beta\rfloor}}\le B_0
\Big\},
\end{align*}
where \(\mathbb N_0=\{0,1,2,\ldots\}\),
\(\partial^\alpha=\partial_1^{\alpha_1}\cdots\partial_k^{\alpha_k}\), and
\(\|\alpha\|_1=\sum_{i=1}^k\alpha_i\). We say that \(f\) is
\(\beta\)-H\"older if \(f\in\c H^\beta([0,1]^k,B_0)\) for \(B_0>0\), where the index \(\beta\) governs the intrinsic smoothness of \(f\).

Smoothness assumptions on the learning target are standard in nonparametric
regression and generative modeling
\citep{jiao2023deep,song2026wasserstein,huang2026theoretical}.
Moving beyond standard configurations, \citet{chen2026error} imposed hierarchical compositional
smoothness on the generator, and \citet{chai2026neural} imposed a related
condition for scalar responses. Under such compositional frameworks, standard H\"older smoothness represents the least structured configuration in which a single component depends on
all inputs. 

In our framework, we apply the standard smoothness condition separately to the sufficient predictor and
the reduced generator. An additional Lipschitz condition
controls how estimation error in the sufficient predictor propagates through their composition.
\begin{assumption}\label{as:sufficient-predictor-smoothness}
For some \(\beta_s,\beta_h,B_s,B_h>0\), each coordinate of \(s^*\) belongs to
\(\c H^{\beta_s}([0,1]^p,B_s)\), each coordinate of \(h^*\) belongs to
\(\c H^{\beta_h}([0,1]^{d+q},B_h)\), and \(h^*\) is Lipschitz in its
\(z\)-input: for all \(z,z'\in[0,1]^d\) and \(e\in[0,1]^q\),
\(\|h^*(z,e)-h^*(z',e)\|\le L_h\|z-z'\|\) for some \(L_h>0\).
\end{assumption}
The Lipschitz requirement is satisfied in several standard regression models. In
nonparametric regression with additive noise \(Y=f(X)+\eps\)
\citep{jiao2023deep}, taking \(z=f(X)\) yields
\(h^*(z,e)=z+e\), which is Lipschitz in \(z\). The location-scale model
\(Y=\mu(X)+\sigma(X)\eps\), with \(\eps\indep X\) and \(\sigma>0\), provides
another example by setting \(s^*(X)=(\mu(X),\sigma(X))\) and
\(h^*(z,e)=z_1+z_2e\).  Notably, 
\citet{tang2026belted} imposed a stronger condition, requiring that all target functions
are Lipschitz across all inputs.

Following approximation theory literature \citep{song2026wasserstein,huang2026theoretical}, we assume that network architectures are sufficiently wide to approximate the target belt \(s^*\) and the reduced 
generator \(h^*\) under fixed envelope constants. Hereafter, \(\asymp\) denotes equality up to positive constants depending
only on the fixed dimensions and smoothness parameters. 
\begin{assumption}\label{as:BeltedEngression-envelope}
\(\bar B_s\ge1\) and \(\bar B_h\ge2B_Y\).
\end{assumption}
\begin{assumption}\label{as:BeltedEngression-architecture}
For \(w_s,w_h\in(0,1/2)\), take
\(N_s=\lceil n^{w_s}/\log^2n\rceil\),
\(N_h=\lceil n^{w_h}/\log^2n\rceil\), and
\(M_s=M_h=\lceil\log n\rceil\). Suppose
\(W_s\asymp N_s\log N_s\), \(D_s\asymp M_s\log M_s\),
\(W_h\asymp N_h\log N_h\), and \(D_h\asymp M_h\log M_h\).
\end{assumption}
The tuning parameters \(N_s, N_h, M_s\) and \( M_h\) govern the approximation capacity of the
neural network class. They grow with the sample size \(n\) and are determined by the exponents \(w_s\) and \(w_h\), which will be formally defined in the subsequent analysis. 
The exact architecture constants characterizing $W_s, D_s$ with respect to \(N_s, M_s\) and \(W_h, D_h\) with respect to \(N_h, M_h\)
are provided in Supplementary
Condition~S.1, while the sizes \(S_s\) and \(S_h\) denote the total parameter counts of the
resulting fully connected architectures.

For real numbers \(a\) and \(b\), write \(a\wedge b=\min\{a,b\}\) and
\(a\vee b=\max\{a,b\}\).
We define the training sample combined with generated noise as
\(
\c S=\{(X_i,Y_i,\eta_{i1},\ldots,\eta_{im}):i=1,\ldots,n\},
\)
and let \(\E_{\c S}\) denote the average over \(\c S\).
Throughout our rate derivations, \(\widetilde O\) suppresses factors that are
polylogarithmic in \(n\) and \(m\), with constants independent of \(n\) and \(m\).
\begin{proposition}\label{prop:BeltedEngression-mn}
Suppose Assumptions~\ref{as:sdr-borel}, \ref{as:bounded},
\ref{as:sufficient-predictor-smoothness},
\ref{as:BeltedEngression-envelope}, and
\ref{as:BeltedEngression-architecture} hold, with the exact architecture
constants in Supplementary Condition~S.1. Then
\[
\begin{aligned}
\E_{\c S}[\cL(\hat s,\hat h)-\cL(s^*,h^*)]
=\widetilde O\!\Big(&\underbrace{
n^{-2\beta_sw_s/p}+n^{-2\beta_hw_h/(d+q)}
}_{\text{approximation error}}\\
&+\underbrace{
n^{2w_s-1}+n^{2w_h-1}
}_{\text{stochastic error}}
+\underbrace{
m^{-1/2}n^{w_s-1/2}+m^{-1/2}n^{w_h-1/2}
}_{\text{Monte Carlo error}}
\Big).
\end{aligned}
\]
\end{proposition}
\begin{remark}\label{rem:mn bd for Bengression}
  The bound in Proposition \ref{prop:BeltedEngression-mn} separates the approximation, stochastic, and Monte Carlo errors for both
the belt and the reduced generator, paralleling the decompositions for 
Engression in \citet{chen2026error}. 
However, a pivotal theoretical innovation of our work is the localized treatment of the stochastic error under standard
    H\"older smoothness conditions. {As established in Supplementary Lemma~S.5, the stochastic error satisfies a Bernstein-type inequality \citep{bartlett2006empirical}. Exploiting this property, Supplementary Lemma~S.6 localizes the analysis to near-optimal candidates \citep{bartlett2005local} and yields a sharper stochastic-error bound.}  In contrast, existing Engression analyses \citep{huang2026theoretical,chen2026error} rely on uniform control over the entire network class, resulting in looser bounds and suboptimal rate exponents.
\end{remark}

While \citet{chai2026neural} provided a fine-grained analysis requiring the Monte Carlo simulation budget to grow rapidly to render its error negligible (that is, $m$ noises are drawn from \(m_1\)
batches of \(m_2\) noises, and  \(m_1m_2\) are required to grow quickly for the Monte Carlo error to be negligible), \citet{chen2026error} and \citet{huang2026theoretical} reported polynomial rates that are independent of \(m\) and valid for \(m\ge2\). Our Theorem \ref{thm:BeltedEngression-rate} optimizes \(w_s\) and \(w_h\) to yield a unified convergence rate for Belted Engression across varying simulation budgets.

\begin{theorem}[Rate of Belted Engression]\label{thm:BeltedEngression-rate}
Under the conditions of Proposition~\ref{prop:BeltedEngression-mn}, suppose
that \(m\asymp n^\kappa\) for \(\kappa\ge0\).
\begin{enumerate}[(1)]
\item Under the oracle choices of \(w_s\) and \(w_h\), there is a threshold
\(\kappa_b^*>0\), depending only on \(p,d,q,\beta_s\), and \(\beta_h\), such
that
\begin{equation*}
\E_{\c S}[\cL(\hat s,\hat h)-\cL(s^*,h^*)]
=
\begin{cases}
\widetilde O\!\left(
n^{-(1+\kappa)\left\{
\frac{\beta_s}{2\beta_s+p}
\wedge
\frac{\beta_h}{2\beta_h+d+q}
\right\}}
\right),
&\mathrm{ if } \ \displaystyle
0\le\kappa<\kappa_b^*,\\[5mm]
\widetilde O\!\left(
n^{-\left\{
\frac{\beta_s}{\beta_s+p}
\wedge
\frac{\beta_h}{\beta_h+d+q}
\right\}}
\right),
&\mathrm{ if } \ \displaystyle
\kappa\ge\kappa_b^*.
\end{cases}
\end{equation*}
\item Moreover, if every coordinate of \(s^*\) is \(\beta\)-H\"older for every
\(\beta>0\), then the oracle choices of \(w_s\) and \(w_h\) yield
\begin{equation*}
\E_{\c S}[\cL(\hat s,\hat h)-\cL(s^*,h^*)]
=
\begin{cases}
\widetilde O\!\left(
n^{-\frac{(1+\kappa)\beta_h}{2\beta_h+d+q}}
\right),
&\mathrm{ if } \ \displaystyle
0\le\kappa<\frac{\beta_h}{\beta_h+d+q},\\[4mm]
\widetilde O\!\left(
n^{-\frac{\beta_h}{\beta_h+d+q}}
\right),
&\mathrm{ if } \ \displaystyle
\kappa\ge\frac{\beta_h}{\beta_h+d+q}.
\end{cases}
\end{equation*}
\end{enumerate}
\end{theorem}
The oracle choices of \(w_s\) and \(w_h\) used in both parts, along with
the threshold \(\kappa_b^*\), are detailed in Supplementary
Condition~S.3.
The two cases effectively distinguish the simulation-limited regime from the saturation threshold, beyond which increasing \(m\) yields no further improvement to the polynomial rate. In either
regime, the final convergence exponent is determined by the slower of the two tasks: learning \(s^*\) and learning \(h^*\).

Part (2) of Theorem \ref{thm:BeltedEngression-rate} formalizes the statistical benefit of SDR for conditional
generation. Engression learns directly on \((X,\eta)\), meaning its standard H\"older
upper bound deteriorates with \(p+q\) \citep{huang2026theoretical}. However, when \(s^*\) is arbitrarily
smooth   
{(that is,  \(s^*\) is \(\beta\)-H\"older for every
\(\beta>0\))}, the complexity of learning the belt vanishes from the final rate, leaving a bound governed by
\(d+q\) and \(\beta_h\). Thus, embedding the belt alters the dimensionality governing the statistical bound, ensuring the rate scales only with the intrinsic sufficient dimension rather than the ambient space $p$.

\label{rem:arbitrarily-smooth-sdr}
Furthermore, Part (2) of Theorem \ref{thm:BeltedEngression-rate} covers classical structures in which \(s^*\) is a linear
projection or coordinate subvector, as seen in sliced inverse regression
\citep{li1991sliced}, generalized linear models \citep{nelder1972generalized}, single-index models 
\citep{lin2007identifiability} and feature screening \citep{fan2008sure}. This regime also covers classical nonlinear parametric models whose
conditional law depends on $X$ through an analytic regression function,
including polynomial, exponential, logistic-growth 
models \citep{davidian2003nonlinear}.

\begin{remark}[Comparison with existing SDR methods.]
The fitted belt \(\hat s\) serves as the learned counterpart to the sufficient
predictor \(s^*\), which is the central target of SDR. While classical sliced inverse
regression and its variants \citep{li1991sliced,li2017nonlinear} can estimate such
predictors, { they often incur substantial computational costs as the sample size increases}, as noted by
\citet{tang2026belted}. The BENN architecture \citep{tang2026belted} improves this scalability, but strictly requires a user-specified ensemble of response transformations and fails to directly generate conditional samples. Alternatively, GenSDR \citep{xu2025conditional}
combines generative modeling and SDR through a flow-based construction. However, GenSDR relies heavily on the regularity of an induced velocity-field ordinary differential
equation, and its theoretical convergence rate of \(n^{-1/(2+p\vee(r+2))}\) also deteriorates as the
response dimension \(r\) grows. In contrast, Belted Engression jointly estimates the sufficient predictor and performs conditional generation via a highly efficient forward pass, achieving a
polynomial convergence exponent that is independent of the response dimension \(r\). 
\end{remark}

\subsection{Convergence Rates for the Engression Baseline}
We next {turn to the convergence rate of the Engression baseline by applying the localized analysis underlying
Proposition~\ref{prop:BeltedEngression-mn}, specifically the Bernstein-type stochastic control}. This approach yields a
unified finite-\((n,m)\) convergence rate that provides a sharper bound than existing results under standard H\"older smoothness when the simulation budget $m$ grows.

We begin by characterizing the smoothness of the unstructured generator \(g^*\) in \eqref{eq:Engression}, analogous to the conditions placed on the sufficient predictor in Assumption~\ref{as:sufficient-predictor-smoothness}.
\begin{assumption}\label{as:betag}
Every coordinate of \(g^*\) belongs to
\(\c H^{\beta_g}([0,1]^{p+q},B_g)\) for some \(\beta_g,B_g>0\).
\end{assumption}

To approximate \(g^*\), we {use}
\(
\c G_{NN}=\c F_{NN}^{p+q,r}(W_g,D_g,S_g,\bar B_g)
\) as the corresponding neural network space, {where the meanings of the parameters $W_g, D_g, S_g, \bar B_g$ are defined in Section \ref{sec:preliminary}.
The next two assumptions impose a condition on $\bar B \lo g$ and   rates on   $W_g, D_g, S_g, \bar B_g$. The exact specifications of the architecture
constants are detailed in Supplementary
Condition~S.1. Hereafter, for any $w \in (0, 1/2)$, let $N_g (w) = \lceil n^w/\log^2 n\rceil$ and $M_g (w) =\lceil\log n\rceil $. }

\begin{assumption}\label{as:engression-envelope}
The fixed envelope constant satisfies \(\bar B_g\ge2B_Y\).
\end{assumption}

\vspace{-.3in}
{\begin{assumption}\label{as:WD}
For \(w\in(0,1/2)\),   \(W_g\asymp N_g (w)\log N_g (w)\) and \(D_g\asymp M_g (w) \log M_g (w) \).
\end{assumption}}

The empirical objective for the unstructured estimator, acting as the finite-sample analogue to \(\cL(g)\) in
\eqref{eq:population-Engression-risk} is formulated as:
\[
\hL(g)
=
\frac{1}{n}\sum_{i=1}^{n}
\left\{
\frac 1m\sum_{j=1}^{m}\|Y_i-g(X_i,\eta_{ij})\|
-
\frac{1}{2m(m-1)}
\sum_{j\ne j'}
\|g(X_i,\eta_{ij})-g(X_i,\eta_{ij'})\|
\right\}, 
\]
{where $\{\eta \lo {ij}: i = 1, \ldots, n, j = 1, \ldots, m \}$ are i.i.d. random vectors distributed as $\mathrm{Unif}([0,1] \hi q)$. }
We denote its empirical minimizer by
\[
\hat g\in\arg\min_{g\in\c G_{NN}}\hL(g).
\]

\begin{proposition}[Rate of Engression]\label{prop:Engression-rate}
Suppose Assumptions~\ref{as:bounded}, \ref{as:betag},
\ref{as:engression-envelope}, and \ref{as:WD} hold. {Moreover, suppose}
\(m\asymp n^\kappa\) for \(\kappa\ge0\), and define the threshold
\(\kappa_e^*=\beta_g/(\beta_g+p+q)\). {Then, under} the oracle choice of \(w\)
detailed in Supplementary
Condition~S.2,
\[
\E_{\c S}[\cL(\hat g)-\cL(g^*)]
=
\begin{cases}
\widetilde O\!\left(
n^{-\frac{\beta_g(1+\kappa)}{2\beta_g+p+q}}
\right),
&\mathrm{ if } \ 0\le\kappa<\kappa_e^*,\\[2mm]
\widetilde O\!\left(
n^{-\frac{\beta_g}{\beta_g+p+q}}
\right),
&\mathrm{ if } \ \kappa\ge\kappa_e^*.
\end{cases}
\]
\end{proposition}

Proposition~\ref{prop:Engression-rate} provides a unified convergence spectrum across varying
simulation budgets. For a fixed, small \(m\), it recovers the standard nonparametric rate
under H\"older smoothness \citep{jiao2023deep}. As \(m\) grows, the simulation error diminishes, and the overall
rate improves until the observation-sample error becomes dominant, yielding the sharper large-\(m\) bound.

Our unified bound in Proposition~\ref{prop:Engression-rate}  facilitates a direct comparison with the existing literature. Three concurrent studies recently derived convergence rates for neural-network Engression \citep{chai2026neural,huang2026theoretical,chen2026error}, and our Proposition \ref{prop:Engression-rate} complements and extends these findings. Our analysis of the {unbelted} Engression baseline aligns most closely with \citet{huang2026theoretical}, which likewise assumes
standard H\"older smoothness over the full generative mapping. In contrast, 
\citet{chen2026error} assumes a composition of low-dimensional generator
components, and \citet{chai2026neural} imposes a hierarchical structure on the
conditional quantile function. The standard H\"older smoothness serves as the
unstructured specification of these compositional models.  The following remarks explain the distinctions between our guarantees and these existing results.

\begin{remark}[Comparison with \cite{chai2026neural}]
For scalar responses, \citet{chai2026neural} obtained a faster theoretical rate by additionally
assuming that the conditional cumulative distribution function (CDF) is uniformly Lipschitz. However, their results do not readily extend to multivariate responses, and this Lipschitz condition excludes target distributions with discrete atoms, a limitation mathematically demonstrated by the counterexample in \citet{chen2026error}.  Furthermore, \citet{chai2026neural} assumed the simulation budget $m$ diverges to a sufficiently large level, which is functionally equivalent to the large-$m$ regime (i.e., $\kappa \ge \kappa_e^*$ ) in our Proposition~\ref{prop:Engression-rate}, while leaving the small-$m$ regime (i.e., $\kappa < \kappa_e^*$) unexplored. 
\end{remark}

\begin{remark}[Comparison with \citet{huang2026theoretical} and \citet{chen2026error}]
  Under the standard H\"older smoothness assumption, the polynomial rates achieved by
\citet{huang2026theoretical} and \citet{chen2026error} coincide with the \(\kappa=0\) (fixed simulation budget) case
of Proposition~\ref{prop:Engression-rate}. While
\citet{chen2026error} explicitly accounts for finite-\(m\) Monte Carlo error,
their analysis does not derive a faster rate as \(m\) increases. Leveraging our localized Bernstein analysis, our result moves beyond fixed-budget bounds to explicitly quantify the structural improvement obtained from increasing $m$ and yields a sharper rate in the
growing-\(m\) regime.
\end{remark}

\subsection{Comparison of Convergence Rates and Network Sizes}\label{sec:rate-comparison}
Having established finite-sample guarantees for both {belted and unbelted Engressions}     under standard H\"older smoothness,
we will quantify the theoretical advantages induced by our belt design. Specifically, we evaluate how this compress-then-generate paradigm improves the statistical efficiency and architectural complexity of conditional generation.

By Proposition~\ref{prop:minLsh}, the Engression generator \(g^*\) and the factorized representation \((s^*,h^*)\) yield the same conditional
law of \(Y\) given \(X\). 
Since the Engression target is identified only through the conditional law it
generates, the composite map \(\bar g(x,e)=h^*(s^*(x),e)\) serves as a mathematically valid formulation for the
full generator target \(g^*\). The following proposition explicitly bounds the intrinsic smoothness of this composite map.
\begin{proposition}\label{prop:composition-smoothness}
Suppose Assumptions~\ref{as:sdr-borel} and \ref{as:sufficient-predictor-smoothness} hold. Let
\(\bar\beta=\min\{\beta_s,\beta_h\}\). Then each coordinate of the composite
map \(\bar g(x,e) = h^*(s^*(x),e)\) is $\bar\beta$-H\"older.
\end{proposition}
{Based on this} smoothness result {we next establish} a rigorous   theoretical comparison of Engression and Belted Engression by aligning the smoothness index \(\beta_g\) in Assumption~\ref{as:betag} with the composite smoothness bound \(\bar\beta\).
\begin{assumption}\label{as:comparison-beta}
The Engression smoothness index in Assumption~\ref{as:betag} and the smoothness
indices of the belt \(s^*\) and generator \(h^*\) in Assumption~\ref{as:sufficient-predictor-smoothness} {satisfy}
\(\beta_g=\min\{\beta_s,\beta_h\}\).
\end{assumption}

\def\eop{\hfill $\Box$}

While the neural network literature frequently assumes Lipschitz target functions \citep{shen2020deep, tang2026belted}, which corresponds to the restrictive setting of
\(\beta_g=\beta_s=\beta_h=1\), our Assumption~\ref{as:comparison-beta} is more general by permitting heterogeneous smoothness across components. The following example illustrates a concrete setting where Assumption~\ref{as:comparison-beta} holds.
\begin{example}\em
  Suppose \(Y\) depends on \(X\) only through a linear transformation \(W^\top X\), where \(W\in\R^{p\times d}\) has full column rank. One may then naturally set
  \(s^*(x)=W^\top x\). In this scenario, \(\beta_s\) can be taken to be arbitrarily
  large. Supplementary
  Proposition~S.1 implies that the
  exact H\"older exponents of \(g^*(x,e)=h^*(W^\top x,e)\) and \(h^*\) are
  equal. Hence,
  Assumption~\ref{as:comparison-beta} holds. \eop
\end{example}

\begin{theorem}\label{thm:rate-comparison}
Suppose Assumption~\ref{as:comparison-beta} and the conditions of Proposition~\ref{prop:Engression-rate} and
Theorem~\ref{thm:BeltedEngression-rate} hold. Then, the difference in rate exponents is strictly positive, i.e., \(\Delta_\kappa>0\), where
\[
\begin{aligned}
\Delta_\kappa
={}&
\underbrace{
\left[
(1+\kappa)
\left\{
\frac{\beta_s}{2\beta_s+p}
\wedge
\frac{\beta_h}{2\beta_h+d+q}
\right\}
\wedge
\left\{
\frac{\beta_s}{\beta_s+p}
\wedge
\frac{\beta_h}{\beta_h+d+q}
\right\}
\right]
}_{\text{Belted Engression exponent}}\\
&-
\underbrace{
\left\{
\frac{(1+\kappa)\beta_g}{2\beta_g+p+q}
\wedge
\frac{\beta_g}{\beta_g+p+q}
\right\}
}_{\text{Engression exponent}}.
\end{aligned}
\]

\end{theorem}

Theorem \ref{thm:rate-comparison} compares the convergence rate of {unbelted} Engression in Proposition~\ref{prop:Engression-rate} {with} that of Belted Engression in Theorem~\ref{thm:BeltedEngression-rate} and shows that the {latter converges faster}.
{As in the discussion} after Theorem~\ref{thm:BeltedEngression-rate} on Page~\pageref{rem:arbitrarily-smooth-sdr}, for classical structures where every coordinate of \(s^*\) is \(\beta\)-H\"older for every \(\beta>0\),
the smoothness index \(\beta_s\) can be taken to be arbitrarily large. Focusing on the regime
\(\kappa\ge1\), where the simulation budget is sufficiently large, the following corollary gives a simplified characterization of this statistical advantage.
\begin{corollary}\label{cor:rate-comparison-smooth}
Suppose the conditions of Theorem~\ref{thm:rate-comparison} hold {and} every
coordinate of \(s^*\) is \(\beta\)-H\"older for every \(\beta>0\). If 
\(\kappa\ge1\), then
\(
\Delta_\kappa
=
\frac{\beta_h}{\beta_h+d+q}
-
\frac{\beta_h}{\beta_h+p+q}
>0.
\)
\end{corollary}

In view of Corollary \ref{cor:rate-comparison-smooth}, the upper-bound exponent for Belted Engression exceeds the upper-bound exponent for Engression by \(\Delta_\kappa\).  Mathematically, this advantage arises because the belted architecture isolates the generative learning task to the dimension-reduced manifold \((s^*(X),\eta)\in\mathbb R^{d+q}\) rather than the ambient space
\((X,\eta)\in\mathbb R^{p+q}\). Under
Assumption~\ref{as:comparison-beta}, the reduction in effective dimension
ensures that the resulting bound remains sharper after accounting for the statistical cost of estimating the belt \(s^*\).

\begin{remark}[Robustness to over-specification]\label{robustness-overspe}
This theoretical guarantee remains robust even if the true minimal sufficient dimension $d_0$ is unknown. Any specified dimension \(d\ge d_0\) still
gives a sharper rate, provided that \(d<p\) and \(X\indep Y\mid s^*(X)\). This result confirms that over-specification of the bottleneck does not invalidate our comparative advantages.
\end{remark}

Beyond statistical efficiency, we must also consider the architectural complexity required to attain these optimal bounds, measured by the total number of network parameters, namely \(S_s+S_h\) for Belted Engression and \(S_g\) for the Engression baseline. 
Define
\(\rho_g(\kappa)=
\Bigl\{\frac{(p+q)(1+\kappa)}{2\beta_g+p+q}\Bigr\}
\wedge
\Bigl\{\frac{p+q}{\beta_g+p+q}\Bigr\}\),
\(\rho_s(\kappa)=
\Bigl\{\frac{p(1+\kappa)}{2\beta_s+p}\Bigr\}
\wedge
\Bigl\{\frac{p}{\beta_s+p}\Bigr\}\), and
\(\rho_h(\kappa)=
\Bigl\{\frac{(d+q)(1+\kappa)}{2\beta_h+d+q}\Bigr\}
\wedge
\Bigl\{\frac{d+q}{\beta_h+d+q}\Bigr\}\). The following theorem bounds the relative size of these prescribed architectures.
\begin{theorem}[Network size comparison]\label{thm:size-comparison}
Suppose Assumption~\ref{as:comparison-beta} and the conditions of Proposition~\ref{prop:Engression-rate} and
Theorem~\ref{thm:BeltedEngression-rate} hold, and let
\(m\asymp n^\kappa\) for a fixed \(\kappa\ge0\). For some constants \(c,C>0\) independent of \(n,m\), the architectures
prescribed by Proposition~\ref{prop:Engression-rate} and
Theorem~\ref{thm:BeltedEngression-rate} satisfy:
\[S_g\ge c\,n^{\rho_g(\kappa)}/(\log n)^2, \quad \mathrm{and} \
S_s+S_h\le C\{n^{\rho_s(\kappa)}+n^{\rho_h(\kappa)}\},\]
with \(\rho_s(\kappa)<\rho_g(\kappa)\) and
\(\rho_h(\kappa)<\rho_g(\kappa)\). Consequently, we have:
\[
\frac{S_s+S_h}{S_g}\to0, \quad \mathrm{ as } \ n\to\infty.
\]
\end{theorem}
Theorem~\ref{thm:size-comparison} reveals a structural benefit of Belted Engression: despite requiring the joint estimation of two distinct neural networks, their combined parameter-growth exponent is strictly smaller than that of the single unstructured network. As a result, the relative size of the belted architecture vanishes asymptotically.

Analogous to Corollary~\ref{cor:rate-comparison-smooth}, the following corollary provides a simplified characterization of this architectural reduction for arbitrarily smooth sufficient predictors in the regime \(\kappa\ge1\).

\begin{corollary}\label{cor:size-comparison-smooth}
Suppose the conditions of Theorem~\ref{thm:size-comparison} hold. If every
coordinate of \(s^*\) is \(\beta\)-H\"older for every \(\beta>0\) and
\(\kappa\ge1\), then, for some constants
\(c,C>0\) independent of \(n,m\), the architectures prescribed by
Proposition~\ref{prop:Engression-rate} and
Theorem~\ref{thm:BeltedEngression-rate} satisfy:
\[
S_g\ge \frac{c}{(\log n)^2}
n^{(p+q)/(\beta_h+p+q)},
\quad  \mathrm{and} \
S_s+S_h\le C n^{(d+q)/(\beta_h+d+q)}.
\]
Consequently, we have: \((S_s+S_h)/S_g\to0\) as \(n\to\infty\).
\end{corollary}
Theorem \ref{thm:size-comparison} and  Corollary~\ref{cor:size-comparison-smooth} show that Belted Engression’s accelerated statistical rate is not an artifact of deploying a massively over-parameterized model class. To the contrary, it achieves superior statistical convergence while demanding a strictly smaller architecture with an asymptotically vanishing relative total parameter count. Thus, embedding the sufficient representational bottleneck simultaneously reduces the dimensionality of the ambient covariate space and compresses the required architectural complexity.

\section{Simulation Studies}\label{sec:Simulation}
We evaluate the empirical performance of Belted Engression through two complementary lenses. First, we benchmark its conditional distribution estimation against the unstructured Engression baseline as the training sample size increases. Second, we assess the capacity of the learned belt to successfully recover the oracle low-dimensional sufficient predictor.

To ensure architectural robustness, we evaluate two distinct noise-injection strategies for the generative network. In the Single noise-layer design, one auxiliary noise vector is concatenated with the predictor at the initial input layer. In the Multiple noise-layer design, independent auxiliary noise is injected into each hidden layer. For Belted Engression, the noise is introduced into the reduced generator \(h\)
after the sufficient representational belt (see Supplementary
Section~S.3.1 for the corresponding network structure).  Both designs are fully supported by our theoretical guarantees:
the Single noise-layer design inherently operates as a ReLU generator on \((s(X),\eta)\), while the
Multiple noise-layer design can be equivalently analyzed as a ReLU generator on \((s(X),\tilde\eta)\) after
concatenating the finite sequence of independent, layer-wise noises into a single composite noise vector \(\tilde\eta\).

We evaluate both the Engression baseline and our Belted Engression 
 (abbreviated as Bengression in figures and tables) across three different data-generating configurations. In all scenarios, the ambient covariate is 
\(X\sim \operatorname{Unif}([0,2]^{100})\),
and the univariate response is generated via
\(
Y=f\{s(X)+\varepsilon\},
\)
where the coordinates of the intrinsic noise \(\varepsilon\) are independent
\(\operatorname{Unif}(-\sqrt{3},\sqrt{3})\).  We define the specific functional forms for three settings below: \texttt{Sin}: \(s(X)=(\sin(\pi X_1),X_3^2)\) and
\(f(u)=\sum_{j=1}^2 u_j\); \texttt{Std}: \(s(X)=(X_1,\operatorname{sd}(X_1,X_2,X_3))\)
and \(f(u)=\sum_{j=1}^2 u_j\), where \(\operatorname{sd}\) denotes the standard
deviation; \texttt{Cubic}: \(s(X)=(X_1,X_2)\) and
\(f(u)=\sum_{j=1}^2 u_j^3/3\). For the \texttt{Sin} and \texttt{Std} settings, the conditional law \(P_{Y\mid X}\) depends on $X$ only through the scalar sum
\(\sum_{j=1}^2s_j(X)\), so the minimal sufficient dimension is exactly one. By setting
\(d=2\) in our architecture, we deliberately over-specify the belt to empirically validate the theoretical robustness to over-specification discussed in Remark \ref{robustness-overspe} of Section~\ref{sec:rate-comparison}.

Note that the intrinsic data-generating noise \(\varepsilon\) is two-dimensional and differs from the algorithmic
generator noise \(\eta\). For the latter, we over-parameterize the noise dimension to \(q=100\) in the implementation. Following
\citet{shen2025engression}, the Multiple design uses an independent
100-dimensional noise vector at every hidden layer. To ensure a rigorous comparison, we conduct five-fold cross-validation at a baseline sample size of $n=1000$ for every combination of data setting, method, and noise architecture. 

Models are evaluated using the empirical energy score from the training objective, computed via 
100 conditionally generated samples per validation covariate. Our hyperparameter grid searches over
the total depth \(D\), hidden width \(W\), learning rate, training iterations,
and, specifically for Belted Engression, the belt depth \(D_s\); see
Supplementary Table~S.1. To guarantee convergence at each scale, the optimization hyperparameters (i.e., learning rate and training iterations) are tuned separately for each $n$ via the same five-fold cross-validation.

The final evaluation draws 500 independent test covariates and generates 500 conditional samples per test covariate. We assess the conditional generation using three metrics: Mean Squared Error (MSE), computed as
\(
n_{\mathrm{eval}}^{-1}\sum_i\|\hat\mu(X_i)-\mu(X_i)\|^2,
\)
where \(\hat\mu(X_i)\) is the generated conditional mean and \(\mu(X_i)\) is the
Monte Carlo approximation of the true conditional mean; Squared Energy Distance (\(\mathrm{ED}^2\)): the
empirical squared energy-distance statistic comparing the generated conditional samples to the true conditional samples; Empirical Coverage: the proportion of true conditional samples falling within the 95\%  empirical prediction interval (bounded by the 2.5\% and 97.5\% empirical quantiles) of the generated samples at the same covariate value. Lower MSE and \(\mathrm{ED}^2\) indicate superior distributional
accuracy, while the empirical coverage should be compared with the nominal 0.95 level.

\begin{table}[!ht]
\centering
\footnotesize
\setlength{\tabcolsep}{2.5pt}
\renewcommand{\arraystretch}{0.95}
\caption{The conditional generation performance for the \texttt{Sin} setting. Rows compare Belted Engression against the Engression baseline across varying sample sizes ($n$) and noise-layer designs (Single vs. Multiple).
Reported values represent empirical means evaluated over 100 independent replicates, with Monte Carlo standard errors provided in parentheses.}
\label{tab:simulation-sin}
\begin{tabular}{@{}l l cc cc cc@{}}
\toprule
\multirow{2}{*}{\(n\)} & \multirow{2}{*}{Method} & \multicolumn{2}{c}{MSE} & \multicolumn{2}{c}{\(\mathrm{ED}^2\)} & \multicolumn{2}{c}{Empirical Coverage} \\
\cmidrule(lr){3-4}\cmidrule(lr){5-6}\cmidrule(lr){7-8}
& & Single & Multiple & Single & Multiple & Single & Multiple \\
\midrule
	\multirow{2}{*}{1000} & Engression & 1.219(0.009) & 1.096(0.011) & 0.576(0.004) & 0.426(0.004) & 0.684(0.002) & 0.835(0.002) \\
	 & Bengression & 0.556(0.005) & 0.552(0.005) & 0.205(0.002) & 0.202(0.002) & 0.932(0.001) & 0.932(0.001) \\
	\cmidrule(lr){1-8}
	\multirow{2}{*}{5000} & Engression & 0.629(0.006) & 0.433(0.006) & 0.248(0.002) & 0.160(0.002) & 0.898(0.001) & 0.961(0.001) \\
	 & Bengression & 0.340(0.004) & 0.336(0.004) & 0.126(0.001) & 0.124(0.001) & 0.949(0.001) & 0.953(0.001) \\
	\cmidrule(lr){1-8}
	\multirow{2}{*}{10000} & Engression & 0.663(0.006) & 0.476(0.005) & 0.269(0.002) & 0.176(0.002) & 0.883(0.001) & 0.953(0.001) \\
	 & Bengression & 0.214(0.008) & 0.210(0.007) & 0.082(0.003) & 0.080(0.003) & 0.943(0.002) & 0.951(0.002) \\
\bottomrule
\end{tabular}
\end{table}

Tables~\ref{tab:simulation-sin}--\ref{tab:simulation-cubic} report the MSE, \(\mathrm{ED}^2\), and empirical coverage for the Sin, Std, and Cubic configurations, respectively, tracking both Engression and Belted Engression as the training sample size $n$ increases. These results demonstrate that embedding the sufficient representational bottleneck improves conditional distribution learning when the response depends on an underlying low-dimensional sufficient predictor. Across all three
settings and both noise-layer designs, Belted Engression achieves substantial reductions in both MSE and
\(\mathrm{ED}^2\) relative to Engression. Belted Engression tends to effectively mitigate the performance gap between the Single and Multiple
noise-layer designs that plagues standard Engression. For the Engression baseline, the Single noise-layer design consistently underperforms. This empirical pattern suggests that algorithmic noise injected solely at the input layer is difficult for the network to leverage effectively when forced to simultaneously process the full ambient covariate vector. However, once the belt compresses the covariates into a low-dimensional sufficient representation, the Single design becomes competitive with the Multiple design.

\begin{table}[!ht]
\centering
\footnotesize
\setlength{\tabcolsep}{2.5pt}
\renewcommand{\arraystretch}{0.95}
\caption{The conditional generation performance for the \texttt{Std} setting. Rows compare Belted Engression against the Engression baseline across varying sample sizes ($n$) and noise-layer designs (Single vs. Multiple).
Reported values represent empirical means evaluated over 100 independent replicates, with Monte Carlo standard errors provided in parentheses.}
\label{tab:simulation-std}
\begin{tabular}{@{}l l cc cc cc@{}}
\toprule
\multirow{2}{*}{\(n\)} & \multirow{2}{*}{Method} & \multicolumn{2}{c}{MSE} & \multicolumn{2}{c}{\(\mathrm{ED}^2\)} & \multicolumn{2}{c}{Empirical Coverage} \\
\cmidrule(lr){3-4}\cmidrule(lr){5-6}\cmidrule(lr){7-8}
& & Single & Multiple & Single & Multiple & Single & Multiple \\
\midrule
\multirow{2}{*}{1000} & Engression & 0.950(0.008) & 0.694(0.008) & 0.454(0.003) & 0.286(0.003) & 0.739(0.002) & 0.853(0.003) \\
 & Bengression & 0.304(0.005) & 0.308(0.005) & 0.115(0.002) & 0.115(0.002) & 0.938(0.001) & 0.940(0.001) \\
\cmidrule(lr){1-8}
\multirow{2}{*}{5000} & Engression & 0.282(0.004) & 0.165(0.006) & 0.123(0.001) & 0.060(0.002) & 0.905(0.001) & 0.964(0.001) \\
 & Bengression & 0.103(0.002) & 0.103(0.002) & 0.039(0.001) & 0.038(0.001) & 0.953(0.002) & 0.956(0.001) \\
\cmidrule(lr){1-8}
\multirow{2}{*}{10000} & Engression & 0.266(0.003) & 0.204(0.003) & 0.116(0.001) & 0.079(0.001) & 0.908(0.001) & 0.950(0.002) \\
 & Bengression & 0.076(0.001) & 0.076(0.001) & 0.028(0.001) & 0.028(0.000) & 0.957(0.001) & 0.959(0.001) \\
\bottomrule
\end{tabular}
\end{table}

\begin{table}[!ht]
\centering
\footnotesize
\setlength{\tabcolsep}{2.5pt}
\renewcommand{\arraystretch}{0.95}
\caption{The conditional generation performance for the \texttt{Cubic} setting. Rows compare Belted Engression against the Engression baseline across varying sample sizes ($n$) and noise-layer designs (Single vs. Multiple).
Reported values represent empirical means evaluated over 100 independent replicates, with Monte Carlo standard errors provided in parentheses.}
\label{tab:simulation-cubic}
\begin{tabular}{@{}l l cc cc cc@{}}
\toprule
\multirow{2}{*}{\(n\)} & \multirow{2}{*}{Method} & \multicolumn{2}{c}{MSE} & \multicolumn{2}{c}{\(\mathrm{ED}^2\)} & \multicolumn{2}{c}{Empirical Coverage} \\
\cmidrule(lr){3-4}\cmidrule(lr){5-6}\cmidrule(lr){7-8}
& & Single & Multiple & Single & Multiple & Single & Multiple \\
\midrule
	\multirow{2}{*}{1000} & Engression & 5.434(0.056) & 4.966(0.093) & 1.023(0.007) & 0.797(0.010) & 0.763(0.002) & 0.847(0.002) \\
	 & Bengression & 2.150(0.034) & 1.918(0.026) & 0.325(0.004) & 0.349(0.004) & 0.951(0.001) & 0.940(0.001) \\
	\cmidrule(lr){1-8}
	\multirow{2}{*}{5000} & Engression & 1.658(0.030) & 1.667(0.059) & 0.287(0.004) & 0.224(0.006) & 0.934(0.001) & 0.980(0.001) \\
	 & Bengression & 0.640(0.013) & 0.617(0.012) & 0.104(0.002) & 0.105(0.002) & 0.970(0.001) & 0.976(0.001) \\
	\cmidrule(lr){1-8}
	\multirow{2}{*}{10000} & Engression & 1.737(0.024) & 1.867(0.047) & 0.297(0.003) & 0.259(0.004) & 0.922(0.001) & 0.970(0.001) \\
	 & Bengression & 0.416(0.005) & 0.407(0.005) & 0.070(0.001) & 0.072(0.001) & 0.972(0.001) & 0.978(0.000) \\
\bottomrule
\end{tabular}
\end{table}

\begin{figure}[!ht]
\centering
\includegraphics[width=0.90\textwidth]{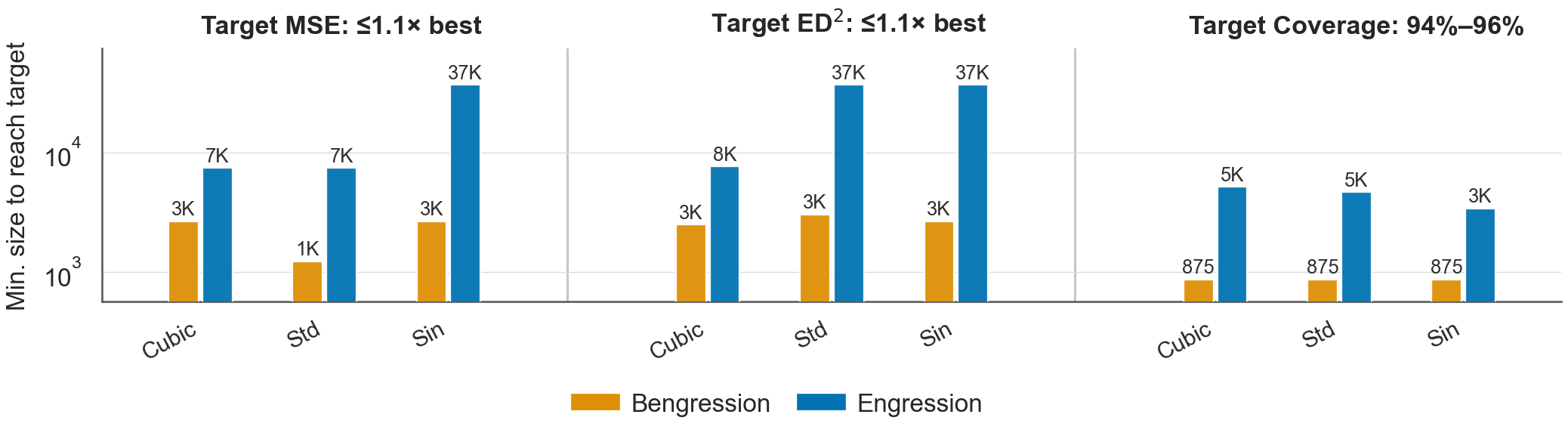}
\caption{Minimum network size required to attain predefined performance thresholds across simulation settings and evaluation metrics.
For MSE and \(\mathrm{ED}^2\), the target is bounded within 10\% of the optimal value achieved for that specific setting, and for empirical coverage, the target interval is  $[0.94, 0.96]$.
}
\label{fig:size-accuracy}
\end{figure}

We also examine the empirical network size required to achieve optimal performance over a fixed architectural grid (see Supplementary Table~S.2). Using the same settings and evaluation metrics, Figure~\ref{fig:size-accuracy} pools the Single and Multiple noise-layer designs to report the minimum parameter count necessary to attain a predefined performance threshold. For MSE and \(\mathrm{ED}^2\), this threshold is defined as falling within 10\% of the best value achieved for that specific setting; for empirical coverage, the target is bounded within the interval $[0.94, 0.96]$. Across different settings and metrics, Belted Engression consistently reaches these target thresholds utilizing a substantially smaller network than Engression. This empirical efficiency validates the architectural complexity bounds mathematically proven in Section~\ref{sec:rate-comparison}.

\begin{figure}[!ht]
\centering
\includegraphics[width=0.90\textwidth]{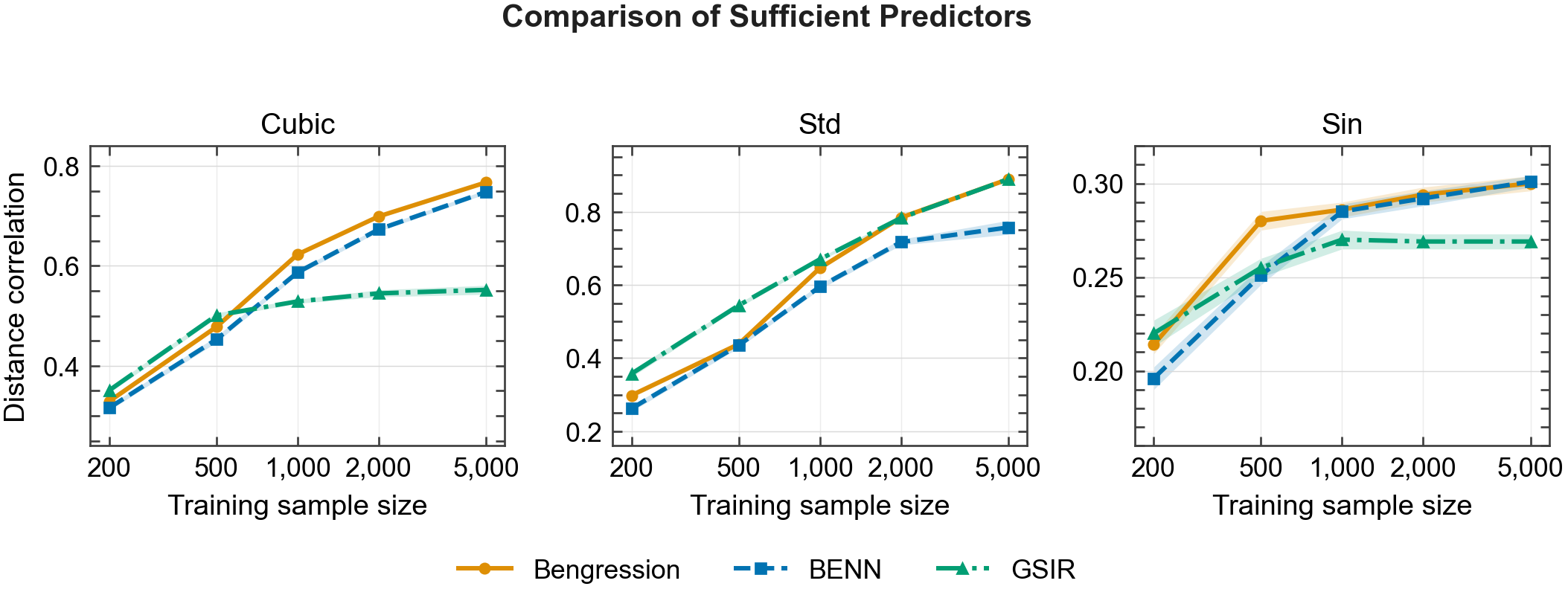}
\caption{Empirical recovery of the sufficient representation across simulation settings and training sample sizes.
Each panel reports the distance correlation between the estimated representation and the oracle sufficient predictor, with shaded ribbons indicating $\pm 1$ Monte Carlo standard error evaluated over 100 independent replicates.
}
\label{fig:sdr-dcor}
\end{figure}

Finally, we compare the learned SDR representation with two established SDR baselines:
GSIR \citep{li2018sufficient},
implemented with a Gaussian kernel, and BENN
\citep{tang2026belted}. For each method, we compute the distance correlation between the estimated representation and the oracle sufficient predictor, where a larger
distance correlation indicates better recovery of the underlying low-dimensional
feature. Recall that the true sufficient predictors are \(\sin(\pi X_1)+X_3^2\) for the \texttt{Sin} setting, 
\(X_1+\operatorname{sd}(X_1,X_2,X_3)\) for the \texttt{Std} setting, and 
\((X_1, X_2)\) for the \texttt{Cubic} setting.
Figure~\ref{fig:sdr-dcor} visualizes these distance correlations across varying sample sizes, evaluated on 500 test samples and averaged over 100 independent replicates. For Belted Engression, we only report the Single noise design, as both noise architectures yield virtually identical SDR recovery. For BENN, we use a Gaussian kernel function class as the ensemble with an ensemble size of \(m_{\text{BENN}}=100\). Empirically, Belted Engression demonstrates highly competitive representation learning: it achieves robust recovery in the \texttt{Sin} setting, matches GSIR while outperforming BENN at larger sample sizes in the \texttt{Std} setting, and strictly overtakes both baselines in the \texttt{Cubic} setting for $n\ge1000$.  These results confirm the efficacy of the belted architecture. The same
objective used to train the conditional generator also extracts an SDR summary aligned with the true low-dimensional structure of the covariates.

\section{Applications}\label{sec:rda}

To demonstrate the practical utility and versatility, we evaluate Belted Engression across two real-world domains: conditional distribution estimation of community crime rates in Section \ref{crime} and conditional image completion on the MNIST dataset in Section \ref{mnist}.

\subsection{Conditional Distribution Estimation of Crime Rates}\label{crime}

We first demonstrate the application of Belted Engression on the Communities and Crime data from the UCI
Machine Learning Repository
(\url{https://archive.ics.uci.edu/dataset/183}). This dataset comprises 102 numeric features capturing community-level demographic,
socioeconomic, housing, and law-enforcement attributes to model the normalized violent-crime rate per population, which is a critical task in the statistical analysis of violent crime \citep[see, e.g.,][]{loeffler2026gun}. The primary statistical task is to estimate the conditional distribution of this crime-rate response given the ambient community covariates.

We use the curated dataset from the R package \texttt{fairml} and impute missing feature values using the training-set medians.
Since the normalized response lies within \([0,1]\), all
models are fitted on the logit scale following a truncation of the response to \([10^{-4},1-10^{-4}]\). The data are partitioned into 1312 training observations and 657 test observations. Model hyperparameters are selected via five-fold cross-validation, employing the energy score on the logit-transformed response as the validation criterion, where the tuning grid and related settings are
reported in Supplementary Table~S.3. Final out-of-sample test metrics are averaged across 10 independent replicate fits.

\begin{table}[ht]
\setlength{\tabcolsep}{2.5pt}
\renewcommand{\arraystretch}{0.95}
\centering
\caption{Out-of-sample predictive performance on the Communities and Crime dataset. Belted Engression is benchmarked against Engression across the Multiple and Single noise-layer designs. Model size denotes the total number of trainable parameters, reported in thousands ($k$) and rounded to one decimal place. Performance metrics include logit-scale MSE, logit-scale $\mathrm{ED}^2$, and empirical 95\% prediction-interval coverage.}
\label{tab:rda-prediction}
\begin{tabular}{@{}c l c c c c@{}}
\toprule
Noise Architecture & Method & Model Size &  MSE (logit) &  \(\mathrm{ED}^2\) (logit) & Coverage \\
\midrule
\multirow{2}{*}{Multiple} & Engression & 45.8k & 214.120 & 1.281 & 0.933 \\
 & Bengression & 16.8k & 191.274 & 1.231 & 0.943 \\
\cmidrule(lr){1-6}
\multirow{2}{*}{Single} & Engression & 11.1k & 207.031 & 1.232 & 0.903 \\
 & Bengression & 5.9k & 191.343 & 1.220 & 0.926 \\
\bottomrule
\end{tabular}
\end{table}

Table~\ref{tab:rda-prediction} reports the
logit-scale MSE, logit-scale \(\mathrm{ED}^2\), and empirical 95\% prediction-interval
coverage. Across both noise-injection architectures, Belted Engression yields lower MSE and \(\mathrm{ED}^2\) compared to Engression, while achieving an empirical coverage closer to the nominal 0.95 level. Belted Engression secures these predictive improvements while demanding fewer trainable parameters. These results confirm that the SDR belt continues to reduce distributional error and improve prediction in real-world observational settings containing over 100 ambient covariates, 
mirroring the structural advantages demonstrated in Section~\ref{sec:Simulation}.

\begin{figure}[!ht]
\centering
\includegraphics[width=0.90\textwidth]{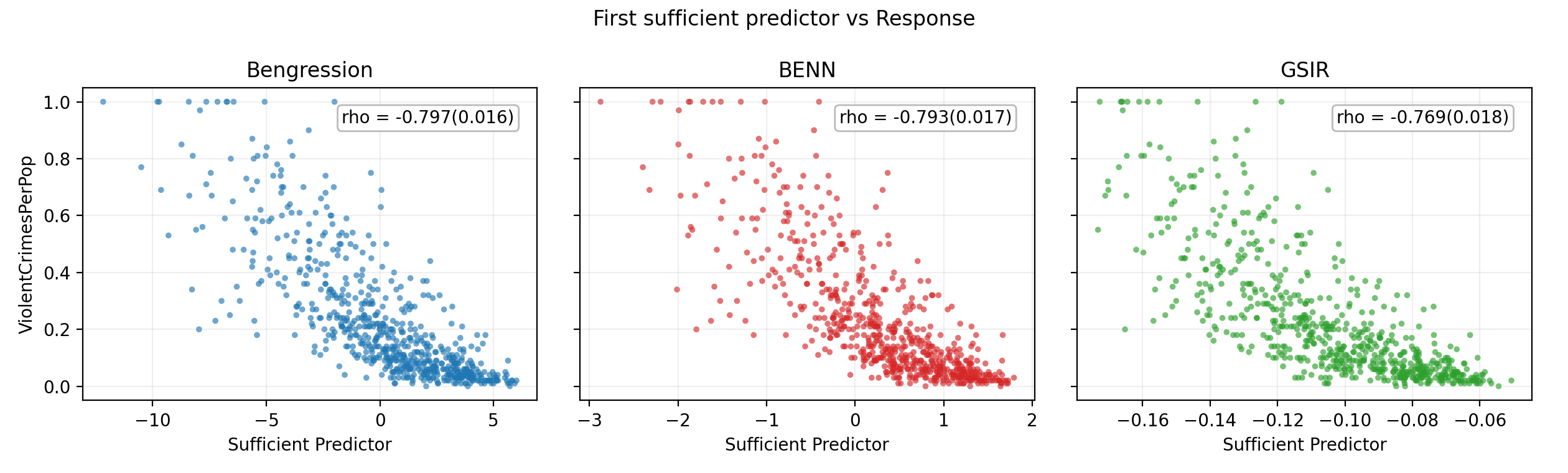}
\caption{Visualization of the one-dimensional SDR representations evaluated on the Communities and Crime test set. Each panel scatters the extracted sufficient predictor against the observed violent-crime response for Belted Engression, BENN, and GSIR.
The displayed \(\rho\) denotes the out-of-sample Pearson correlation between the learned representation and the response, with the corresponding standard error shown in parentheses.}
\label{fig:rda-sdr}
\end{figure}

Figure~\ref{fig:rda-sdr} visualizes the extracted one-dimensional SDR representation evaluated on the
test set for Belted Engression, BENN, and GSIR. To facilitate this visual comparison, both Belted Engression and BENN are configured with a bottleneck dimension of \(d=1\). Each panel plots the fitted one-dimensional representation against the observed violent-crime response. The
reported \(\rho\) value denotes the out-of-sample Pearson correlation on the test set, with the corresponding standard error shown in parentheses. Because a one-dimensional SDR subspace is invariant to orientation, the sign of \(\rho\) is not intrinsically meaningful, but its magnitude quantifies how effectively the learned representation captures the variation in the response. Belted Engression demonstrates an association strength comparable to BENN and strictly superior to GSIR, while simultaneously delivering the superior predictive metrics detailed in Table~\ref{tab:rda-prediction}. Thus, Belted Engression successfully unifies accurate conditional estimation with the extraction of an informative low-dimensional structural summary.

\subsection{Conditional Image Completion on MNIST}\label{mnist}
We examine a conditional sampling task using MNIST handwritten digits. Each image is split into an observed upper half and an unobserved lower half: the covariate \(X\in[0,1]^{392}\) contains the upper 14 rows of the \(28\times28\) image, and the response \(Y\in[0,1]^{392}\)
contains the lower 14 rows. The statistical task is to learn the conditional
distribution of the missing lower half given the observed upper half, with
conditional samples representing possible completions. This setting tests
the utility of the SDR belt: while the ambient pixel space is large, the intrinsic information required for accurate completion is expected to concentrate strictly on low-dimensional digit features.

We train both Engression and Belted Engression on a balanced training sample of 8,000 MNIST
images spanning all 10 digits. For training stability, the target lower-half pixels \(v\) are binarized at 0.2, replacing each with
\(\mathbf 1\{v>0.2\}\). Also, the generator output uses a sigmoid activation
\(\sigma(t)=1/(1+e^{-t})\) to ensure the generated pixels are properly bounded to the image domain. The architectural specifications are provided in Supplementary Table~S.4.

Figure~\ref{fig:mnist-completions} compares the conditionally sampled completions from both methods. The first column of each panel displays the ground-truth image, while subsequent columns present independent conditional samples generated by fixing the upper half. Although both methods learn digit-specific conditional distributions, the completions produced by Belted Engression exhibit greater visual coherence and structural fidelity in the generated lower half.

\begin{center}
\centering
\begin{minipage}{0.32\textwidth}
\centering
\includegraphics[width=\linewidth]{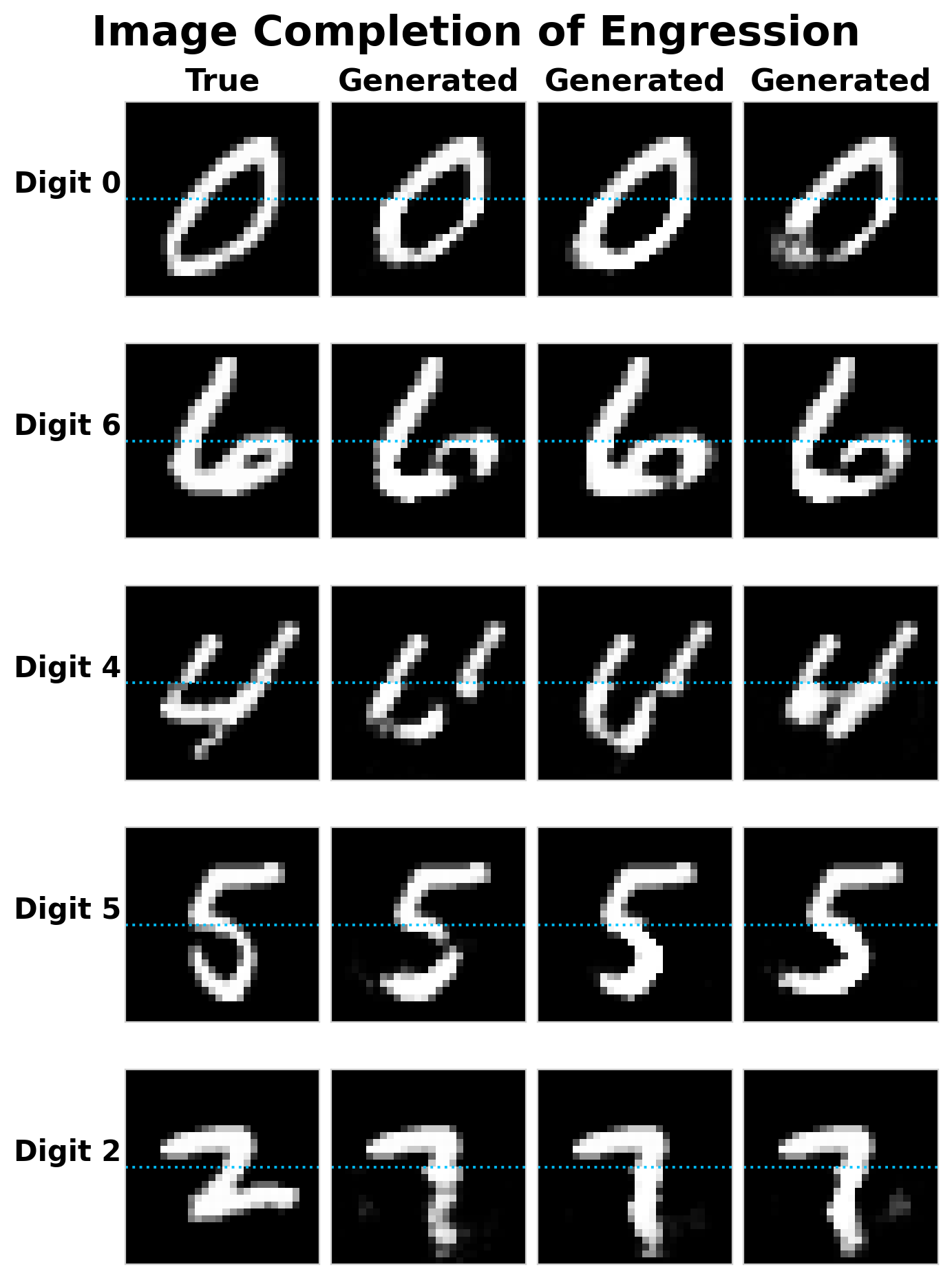}
\end{minipage}\hspace{0.04\textwidth}
\begin{minipage}{0.32\textwidth}
\centering
\includegraphics[width=\linewidth]{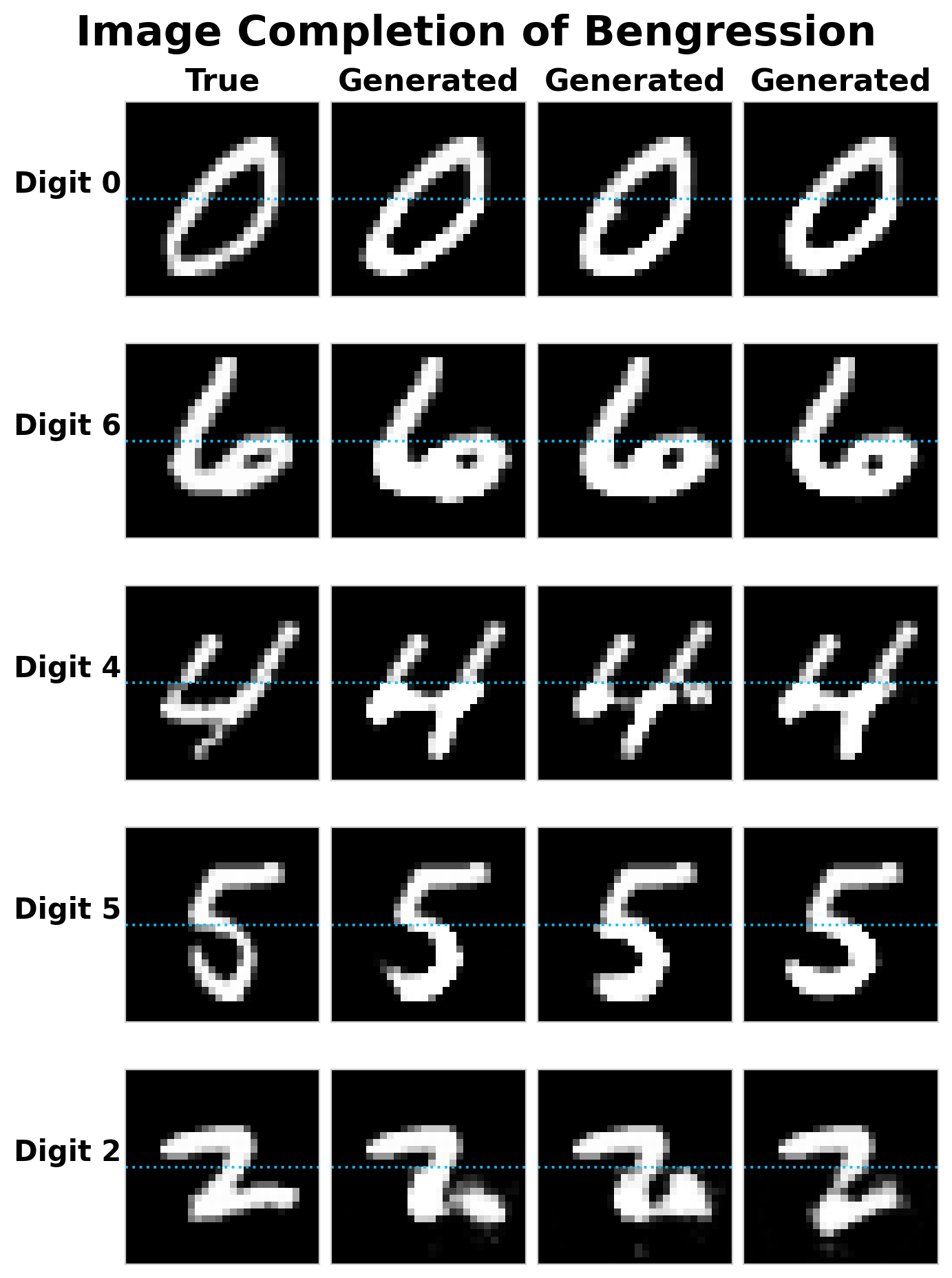}
\end{minipage}
\captionof{figure}{Conditional image completions on the MNIST test set.
In each panel, the first column gives
the ground-truth test image, while subsequent columns give independent conditional samples generated by fixing the same upper half. The dotted horizontal line separates the observed upper-half conditioning covariates from the generated lower-half response.}
\label{fig:mnist-completions}
\end{center}

\begin{figure}[!ht]
\centering
\includegraphics[width=0.5\textwidth]{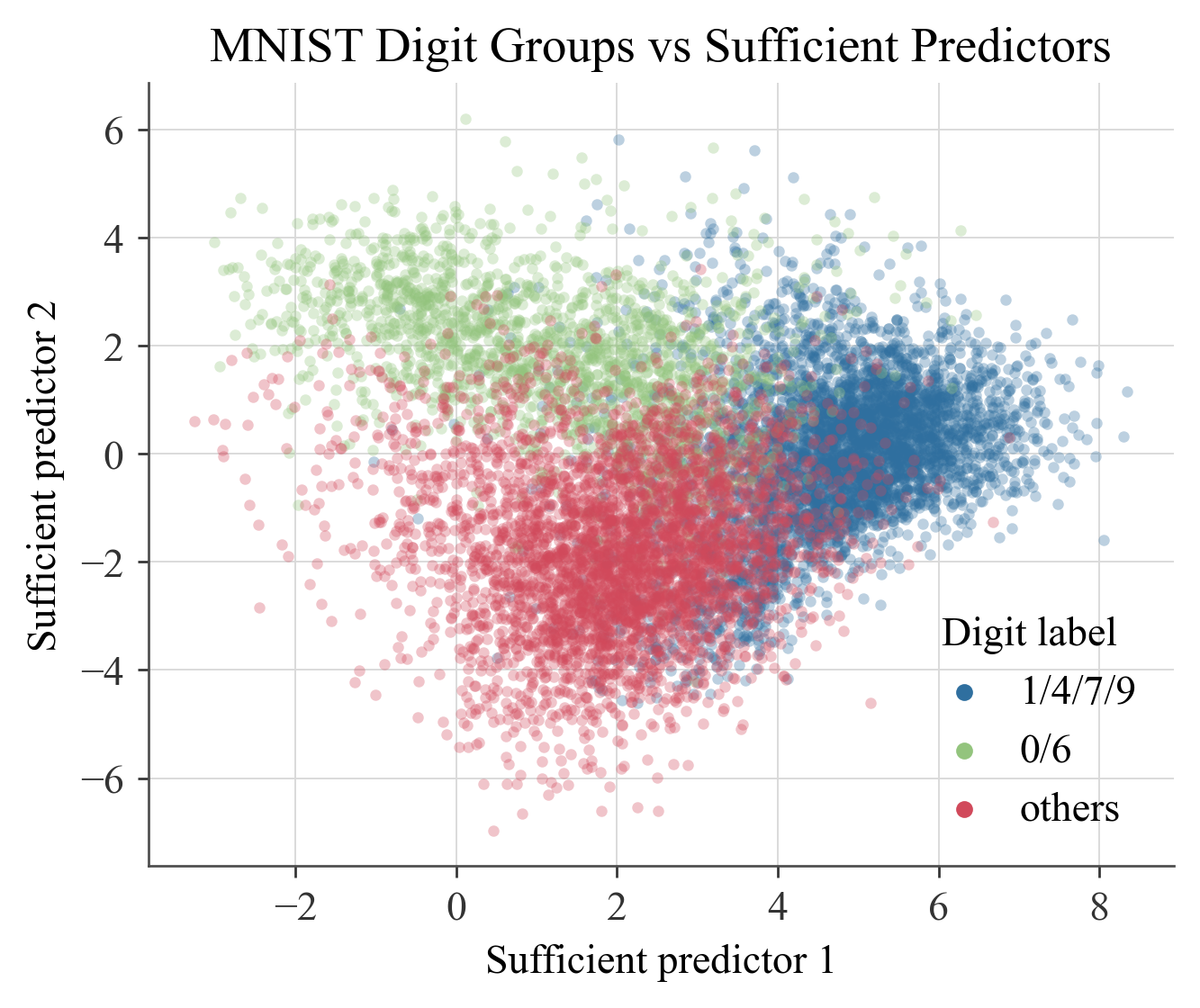}
\caption{Visualization of the SDR extracted by Belted Engression on the MNIST test set. Each point is a test image projected onto two selected coordinates of the learned 10-dimensional SDR bottleneck. Digit labels were withheld during model training and are used only to select these coordinates and color the scatterplot, highlighting the morphological separation between line-like digits \(\{1,4,7,9\}\),
loop-like digits \(\{0,6\}\), and other digits.}
\label{fig:mnist-sdr}
\end{figure}

Beyond generative accuracy, the embedded belt gives an interpretable low-dimensional summary of the
completion task. Figure~\ref{fig:mnist-sdr} plots two specific coordinates from the 10-dimensional SDR bottleneck extracted by Belted Engression on the test set. These coordinates are selected based on their marginal correlations with two coarse digit-shape
groups: line-like digits \(\{1,4,7,9\}\) and loop-like digits \(\{0,6\}\).
The resulting representation separates these groups from the remaining digits, despite the Belted Engression loss function never having access to the categorical digit labels. Thus, the belt learned for conditional generation also organizes the ambient covariate space according to the shape features
driving the completion task. In this complex image-completion scenario, Belted Engression generates conditional samples for a multivariate, 392-dimensional response while concurrently extracting an SDR representation endowed with morphological structure.

\section{Conclusion}\label{sec:discussion}
This paper introduces Belted Engression by integrating a sufficient dimension reduction (SDR) bottleneck into the generative architecture {of the neural network Engression}. Theoretically, we establish that the SDR condition is equivalent to a law-preserving factorization of the conditional distribution, and our localized finite-sample analysis yields unified convergence rates for both methods under fixed and growing network budgets. Under comparable conditions {with the unbelted counterpart}, Belted Engression achieves a sharper upper-bound exponent {for its convergence rate}, and its prescribed architectures are {substantially} smaller, yielding an asymptotically vanishing relative parameter count. Empirically, simulations confirm that our structure accelerates conditional distribution learning and improves SDR recovery with fewer parameters. Finally, our applications to crime rates and MNIST image completion demonstrate that the extracted belt can {effectively} exploit low-dimensional intrinsic geometry in conditional generation without compromising the population target.

{
\setstretch{1}
\bibliographystyle{agsm}
\bibliography{ref}
}
\end{document}